\documentclass{article}

\usepackage{microtype}
\usepackage{graphicx}
\usepackage{subfigure}
\usepackage{booktabs} 

\usepackage{hyperref}

\usepackage[accepted]{icml2024}

\usepackage{amsmath}
\usepackage{amssymb}
\usepackage{mathtools}
\usepackage{amsthm}
\usepackage{wrapfig}
\usepackage{caption}
\usepackage{xcolor}
\usepackage[capitalize,noabbrev]{cleveref}

\theoremstyle{plain}

\theoremstyle{definition}

\theoremstyle{remark}

\usepackage[textsize=tiny]{todonotes}

\icmltitlerunning{Sparse Concept Bottleneck Models: Gumbel Tricks in Contrastive Learning}

\begin{document}

\twocolumn[
\icmltitle{Sparse Concept Bottleneck Models: \\ Gumbel Tricks in Contrastive Learning}



\icmlsetsymbol{equal}{*}

\begin{icmlauthorlist}
\icmlauthor{Andrei Semenov}{MIPT}
\icmlauthor{Vladimir Ivanov}{Innopolis University}
\icmlauthor{Aleksandr Beznosikov}{MIPT}
\icmlauthor{Alexander Gasnikov}{MIPT,Innopolis University}
\end{icmlauthorlist}

\icmlaffiliation{MIPT}{Moscow Institute of Physics and Technology, Moscow, Russia}
\icmlaffiliation{Innopolis University}{Innopolis University, Russia, Innopolis}

\icmlcorrespondingauthor{Andrei Semenov}{semenov.andrei.v@gmail.com}
\icmlcorrespondingauthor{Vladimir Ivanov}{v.ivanov@innopolis.ru}
\icmlcorrespondingauthor{Aleksandr Beznosikov}{beznosikov.an@phystech.edu}
\icmlcorrespondingauthor{Alexander Gasnikov}{gasnikov@yandex.ru}

\icmlkeywords{Machine Learning, ICML}

\vskip 0.3in
]



\printAffiliationsAndNotice{}  

\begin{abstract}
We propose a novel architecture and method of explainable classification with Concept Bottleneck Models (CBMs). While SOTA approaches to Image Classification task work as a black box, there is a growing demand for models that would provide interpreted results. Such a models often learn to predict the distribution over class labels using additional description of this target instances, called concepts. However, existing Bottleneck methods have a number of limitations: their accuracy is  lower than that of a standard model and CBMs require an additional set of concepts to leverage. We provide a framework for creating Concept Bottleneck Model from pre-trained multi-modal encoder and new CLIP-like architectures. By introducing a new type of layers known as Concept Bottleneck Layers, we outline three methods for training them: with $\ell_1$-loss, contrastive loss and loss function based on Gumbel-Softmax distribution (Sparse-CBM), while final FC layer is still trained with Cross-Entropy. We show a significant increase in accuracy using sparse hidden layers in CLIP-based bottleneck models. Which means that sparse representation of concepts activation vector is meaningful in Concept Bottleneck Models. Moreover, with our Concept Matrix Search algorithm we can improve CLIP predictions on complex datasets without any additional training or fine-tuning. The code is available at: \href{https://github.com/Andron00e/SparseCBM}{https://github.com/Andron00e/SparseCBM}.
\end{abstract}
\vspace{-0.3cm}
\section{Introduction}
\label{submission}

In recent years SOTA \cite{yu2023noisynn} approaches to image classification have reached a significant accuracy performance. The main benchmark for checking the quality of such models is ImageNet LSVRC \cite{russakovsky2015imagenet}, and the most advanced models are trained from several weeks to months at large-scale datasets like JFT-300M \cite{sun2017revisiting} and ImageNet-21K \cite{5206848} on many GPU machines. While classic modern solutions, such as ViT \cite{dosovitskiy2021image} and other Vision Transformer models outperform ResNet-based \cite{he2015deep} solutions, although substantially less computational resources are required for pre-training, they still work like a black box. Given a mini-batch of images and necessary labels as the input they produce a ready-made distribution over the class labels. In this case, the goal of Concept Bottleneck Models \cite{koh2020concept} approach is to design models that would answer the questions like: "Why did you choose this particular class?", "Based on what results did you predict it?". This approach is supposed to be extremely useful in domain where explanations are crucial, for instance, in medical applications. 
\begin{figure}[t]
\begin{center}
\centerline{
\includegraphics[width=0.55\columnwidth]{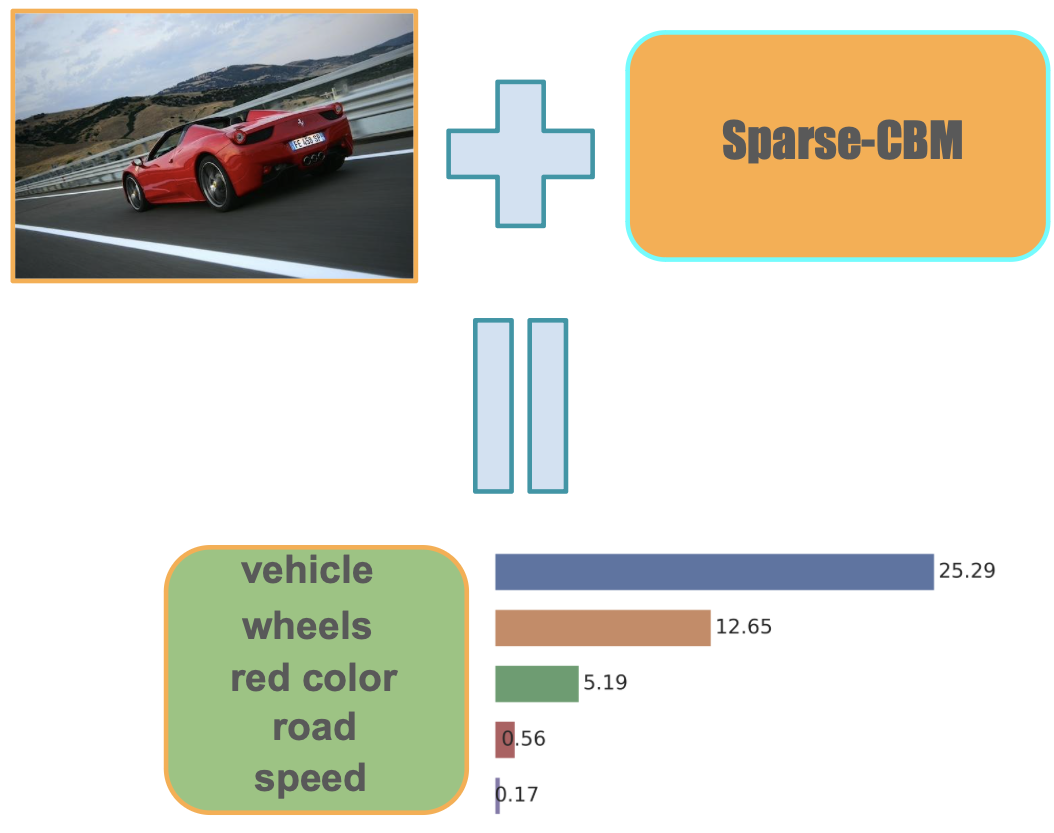}}
\caption{Example of concepts extraction with Sparse-CBM.}
\label{fig:opening}
\end{center}
\vskip -0.4in
\end{figure}
\\
\\
We study models that, firstly, predict an intermediate set of human-understandable concepts and, then, leverage this to predict the final label supported by selected concepts. Building of interpretable neural networks has become more popular with presenting OpenAI CLIP \cite{radford2021learning} model and development of Contrastive Learning techniques \cite{aljundi2022contrastive}. We cannot fail to note the growing multi-modality \cite{reed2022generalist,chen2023sharegpt4v} trend with its promising results with combining several modalities such as text, audio, video to build one model able to solve many downstream tasks. Studies of Concept Bottleneck models and Contrastive Learning also implies multi-modality (between image-text pairs) both in supervised \cite{khosla2021supervised} and unsupervised \cite{gao2022simcse} settings, and even comes in handy for Reinforcement Learning applications \cite{srinivas2020curl}.

\subsection{Contrastive Learning}

Our work is based mostly on Contrastive Representation Learning approaches to classification task, when the main goal is to learn an embedding space so that \textcolor{blue}{the} similar samples make stay closer, while the dissimilar ones are far apart. In case of Contrastive Learning, we aim to learn a joint latent space of image and text embeddings as CLIP, ALIGN \cite{jia2021scaling}, BLIP \cite{li2022blip}, and LiT \cite{zhai2022lit} do.
The softmax-based contrastive loss \cite{1467314} has become a key objective of pre-training such models. However, recent work \cite{zhai2023sigmoid} has shown an ability to outperform previous results with sigmoid loss pre-training, which requires significantly less memory and, thus, enables training of locked-image model from \cite{zhai2022lit} with much bigger batch size. Most of Contrastive Learning methods are applied in data augmentation, but not only for images in computer vision purposes. Prior works \cite{wei2019eda,kobayashi2018contextual,fang2020cert,shen2020simple} have shown how text can be augmented without altering the whole semantics of a sentence.

\subsection{Classification with Concept Bottleneck}\label{sec:bottls}

Introduced by \cite{koh2020concept}, Concept Bottleneck Models became a popular branch in explainable AI. General idea and pipeline of training bottlenecks are quite intuitive: instead of solving a task of interest directly,
let us divide the main problem into smaller parts and retrieve the most meaningful concepts from them. As concepts we can consider any human-understandable information. For instance, when trying to classify a broken arm, orthopedic surgeon looks for special vision patterns, e.g., cracks on the bones, unrealistic crooks of arms etc. CBMs are trained to predict human-understandable concepts at first, and later, to make a final decision based on received concepts. It is worthy that for this very needs end-to-end training is not compulsory, many prior works we refer to in Section \ref{sec:relwork} introduce their frameworks to build CBM from existing pre-trained feature-extractors. What is the advantage of CBM if the accuracy has worsened compared to typical classifier? Lower performance is certainly a key issue in bottleneck NNs, but instead, we try to leverage this parameter to define a trade-off between model's accuracy performance and interpretability. By constraining the model to rely on an intermediate set of concepts and final predictions from a Fully Connected (FC) layer we can: 
\begin{description}
\item \underline{Explain} what information does the model perceive as more/less important for classifying an input.
\item \underline{Understand} why model makes a particular mistake: to determine which concepts were given the wrong priority.
\end{description}
The above mentioned benefits of general Concept Bottleneck Models are followed by key restrictions:
\begin{description}
\item \underline{Concepts preparation:} Training CBMs require a labels marked with additional concepts, which is time consuming and expensive to collect.
\item \underline{Performance:} It is not promising in practice to use models which are lower in accuracy then unrestricted ones.
\item \underline{Editing model:} Common problem raised in Concept Bottleneck Model's approach is unavailability to holistically intervene in model itself. Prior work \cite{yuksekgonul2023posthoc} demonstrates methods of human-guided feedback, while \cite{oikarinen2023labelfree} relies on TCAV\cite{kim2018interpretability} which still pays attention to local model's mistakes in case when CBM is not learned in end-to-end manner. While usually users would like to get pre-trained CLIP-like model and create a bottleneck one above it. Therefore, this question still remains open.
\end{description}

\subsection{Contributions}

In this paper, we contribute to a better understanding of latter approaches to image classification with CBMs. In particular, we discover a  capabilities of CBMs trained in self-supervised manner and reveal a significant growth of performance with models that generate a sparse inner representation. We seek to uncover previously hidden capabilities of raw CLIP-like models with introduced in Section \ref{sec:cms}  "zero-training" algorithm with proposed in Section \ref{sec:framework} framework. Only by leveraging dot-product scores of CLIP we can make our model more interpretable in sense of relying on intermediate set of concepts. Moreover, all our approaches support manual concept generation. We formulate our further framework as a special case of contrastive fine-tuning, when pre-trained models are then fine-tuned via contrastive objective functions \cite{weng2021contrastive}.  

Instead of developing a specific-case architectures, we seek contrastive fine-tuning algorithms that can work with general framework of creating Concept Bottleneck from a pre-trained multi-modal encoder. Furthermore, our approach relies on some previously introduced frameworks with a change in the objectives that hidden layers are trained with. Notably, since a backbone model's structure is well-parallelizable, our CBMs can be maintained by vertical or data parallelism with relative ease. This allows us to design a bottleneck that can solve its task more effectively.

We summarize the main contributions of our work as follows:

\textbf{(\underline{Contribution 1})} We formally define an algorithm to improve the accuracy of CLIP and at the same time make model more interpretable. We also provide a CLIP latent space analysis and how concepts affect it. 

\textbf{(\underline{Contribution 2})} Based on our analysis, we formulate a framework for building CBM: novel architecture and training method which includes a  contrastive fine-tuning of additional layers. We implement an automatic concept set generation based on provided dataset and contrastive variants of our loss functions, thus, our framework facilitates the training of new layers.

\textbf{(\underline{Contribution 3})} We demonstrate the efficiency of Sparse-CBM (see Section \ref{sec:sparsecbm}) method by running our architecture on ImageNet \cite{russakovsky2015imagenet}, CUB-200 \cite{wah_branson_welinder_perona_belongie_2011}, Places365 \cite{7968387}, CIFAR100 and CIFAR10 \cite{krizhevsky2009learning} datasets. As a result, Sparse-CBM outperforms a prior Laber-free \cite{oikarinen2023labelfree} work at several of them and achieves an accuracy of 80.02\% on CUB-200.

\section{Related Work}\label{sec:relwork}

Approaches to Image Classification with Concept Bottleneck models is a rapidly developing field due to mentioned in Section \ref{sec:bottls} benefits of their capabilities to be interpreted, then, many sufficient methods has already been introduced. Early works of \cite{koh2020concept,losch2019interpretability,Marcos_2020_ACCV} proposes to train CNN \cite{5537907} and create an additional layer before the last fully connected layer where each neuron corresponds to a human interpretable concept. Since bottleneck models suffer from the need to prepare a set of concepts, finding an efficient method to create a set of supported concepts is a necessary issue to solve. Prior works on Label-free, Post-hoc CBM and LaBo \cite{oikarinen2023labelfree,yuksekgonul2023posthoc,yang2023language} establish a trend towards creating frameworks (how to convert an existing model to concept bottleneck one) instead of learning models from scratch. They also show methods of creating a set of sufficient concepts and introduce new metrics in classification with CBM. Since a well-prepared set of concepts is a key ingredient of the CBM problem, \cite{schwalbe2022concept} suggest a general survey on methods that leverage concepts embeddings. Among other things, the authors of \cite{oikarinen2023labelfree} train model with two algorithms: classifier's head with GLM-SAGA \cite{pmlr-v139-wong21b} solver and bottleneck layers with variants of cosine similarity, which is close to our pipeline. \cite{menon2022visual} report to a method which do not require training of the basic model, instead, they ask CLIP for specific pattern and find a final class with proposed formula, while \cite{oikarinen2023clipdissect} proposes a way to describe neurons of CLIP hidden layers. \cite{gao2022pyramidclip} suggests building an input pyramid with different semantic levels for each modality and aligning visual and linguistic elements in a hierarchy, which is convenient for object detection. \cite{pmlr-v139-wong21b} introduces an idea to keep final FC layer sparse, thus, making it more interpretable. \cite{kazmierczak2023clipqda} intervenes into the CLIP latent space and shows it can be effectively modeled as a mixture Gaussians. We also mention that in recent work \cite{Alukaev_2023} proposes to use Gumbel sigmoid after a feature extractors (for concepts and images) that produced a sparse concept representation. The idea of sparsity also finds its continuation in \cite{panousis2023discover}, where authors introduce a new type of Local Winner-Takes-All \cite{panousis2019nonparametric} based layers, that arise activation sparsity. Prior works on sparsity are also similar to ours but we, in turn, create a contrastive loss variants for CBM training.

\vspace{-0.3cm}
\section{Preliminaries}\label{sec:prel}
\vspace{-0.2cm}

In this section, we provide all the notation necessary to introduce our methods of training Concept Bottleneck models.
\vskip 0.1in
\paragraph{Notation.}\label{sec:notation}\vspace{-0.2cm} We primarily work with OpenAI CLIP, it consists of two encoders, one for images and another for text. These encoders allow us to obtain a vector representation for both types of data in a multidimensional space of the same dimension (typically 512). Therefore, we provide the following notation. We refer to $f_{\mathrm{T}}(t, \theta)$ as output of text encoder $f_{\mathrm{T}}$, i.e., text embeddings of a batch of text examples $t$. By $\theta$ we mean the parameters of the text encoder, if it is not trainable, we omit these parameters. Also, since in most experiments we tune only a small number of weights of the original CLIP model (see \cref{tab:backbone_nets}), and only fully train new embedded layers, for simplicity we specify the weights of $\theta$ with the same symbol for both encoders. Given a mini-batch of images $x$ we use $f_{\mathrm{I}}(x, \theta)$ for image encoder $f_{\mathrm{I}}$ output, i.e., embeddings of input images, each of dimension of 512 for the main configurations of the CLIP model. $\langle , \rangle$ refers to scalar (dot) product, by $\times$ we denote a Cartesian product of two sets. For vectors, $\|z\|$ is the norm on the vector space which $z$ belongs to. As a rule, we mean by this designation the $2$-norm for vectors from $\mathbb{R}^{n}$ or Frobenius norm for matrices from $\mathbb{R}^{m\times n}$, until otherwise stated.

\section{Problem Statement and Main
Results}\label{sec:probl}

In this section, we present our algorithm and a framework for building CBM. The algorithm presented in Section \ref{sec:cms} provides an increase in the accuracy of CLIP along with its interpretability, whereas the framework (see Section \ref{sec:framework}) ensure a contrastive fine-tuning of the backbone bimodal encoder.

\subsection{Problem Statement}

Along with the Image Classification task, we formalize the classification with \textbf{Concept Bottleneck}. Given the image and text encoders, we intend to learn an additional projection from the embedding space provided by the encoders into the vector space corresponding to the representation of image concepts, and, after that, learn a final projection to obtain class labels probabilities. We intend to learn the following mapping: $x\times t \to \omega \to l$, according to \textit{Notation} \ref{sec:notation}: $x$ stands for image domain, $t$ means a textual concepts, while $\omega$ and $l$ are concepts representation and class labels correspondingly. Which is different from standard $x\to \omega\to l$ scheme because we are able to create concepts $t$ in automated way.

\subsection{Concept Matrix Search algorithm}\label{sec:cms}

We introduce the Concept Matrix Search (CMS) algorithm, which uses CLIP's capabilities to represent both images and texts data in joint latent space of the same dimension. At first, we formulate a hypothesis about our data, taking into account how the main CLIP model is trained.

Let us define a set of image embeddings as $\mathrm{I} = \{i_1, \dots, i_{|\mathrm{I}|}\}$, where $i_{\mathrm{k}} = f_{\mathrm{I}}(x_{\mathrm{k}}, \theta)$. And as $\mathrm{D} = \{d_1, \dots, d_{|\mathrm{D}|}\}$ we define a set of concepts embeddings, where $d_\mathrm{j} = f_{\mathrm{T}}(\mathrm{concept}_{\mathrm{j}}, \theta)$, while $\mathrm{C} = \{c_1, \dots, c_{|\mathrm{C}|}\}$ is a set of class embeddings, where class labels embeddings are similarly expressed in terms of $f_{\mathrm{T}}$. We also define two matrices for which our model will compute pairwise image-texts similarity:

Image-Concept matrix (V-matrix): $\mathcal{V}\in \mathbb{R}^{|\mathrm{I}|\times|\mathrm{D}|}$, such that $\mathcal{V}_{\mathrm{kl}} =\langle i_\mathrm{k}, d_\mathrm{l}\rangle$.

Class-Concept matrix (T-matrix): $\mathcal{T}\in\mathbb{R}^{|\mathrm{C}|\times|\mathrm{D}|}$, such that $\mathcal{T}_{\mathrm{kl}} = \langle c_\mathrm{k}, d_\mathrm{l} \rangle$.

Now we can formulate the \textit{hypothesis} underlying the CMS.
\paragraph{Hypothesis.}
For $f_{\mathrm{I}}(x,\theta)$ and $f_{\mathrm{T}}(t,\theta)$ denote a cosine similarity as:
\[ 
\cos(f_{\mathrm{I}}(x, \theta), f_{\mathrm{T}}(t, \theta)) = \frac{\langle f_{\mathrm{I}}(x, \theta), f_{\mathrm{T}}(t, \theta) \rangle}{\|f_{\mathrm{I}}(x, \theta)\| \|f_{\mathrm{T}}(t, \theta)\|}_.
\]
Then,
\begin{quote}
\textit{Given a matrices} $\mathcal{V}$, $\mathcal{T}$ \textit{and function} $\phi(\mathrm{m})\colon \mathbb{N}^{|\mathrm{I}|} \to \mathbb{N}^{|\mathrm{C}|}$ \textit{that maps the embedding index of an image in} $\mathcal{V}$ \textit{to the index of its class in} $\mathcal{T}$\textit{, we presume:}
\[
\forall \mathrm{j}\neq \mathrm{k} \hookrightarrow \cos(\mathcal{V}_{\mathrm{k},\cdot}^\top, \mathcal{T}_{\phi(\mathrm{k}),\cdot}^\top) \geq \cos(\mathcal{V}_{\mathrm{k},\cdot}^\top, \mathcal{T}_{\phi(\mathrm{j}),\cdot}^\top)_,
\]
\textit{where} $\mathcal{V}_{\mathrm{k},\cdot}$, $\mathcal{T}_{\mathrm{k},\cdot}$ \textit{are the k-th rows of} $\mathcal{V}$, $\mathcal{T}$ \textit{respectively.}
\end{quote}
Simply, the \textit{hypothesis} states that for all possible classes, the vector of similarities between the true class and all concepts should be the closest to the vector of similarities of the image of this very class and all concepts. $\mathcal{V}_{\mathrm{k},\cdot}^\top$ vector is obtained by computing the cosine similarity between $i_{\mathrm{k}}$ embedding
and each of the concept's embeddings in $\mathrm{D}$.
\vskip 0.1in
\paragraph{CMS algorithm.} Having formulated the \textit{hypothesis}, we provide an effective method for testing it using \cref{Alg:CMS}. In order to reduce computational costs and memory, we propose in \cref{Alg:CMS} to process the matrix $\mathcal{V}$ in batches of batch size $|\mathrm{B}|<|\mathrm{I}|$. Note that actually we do not run two loops, but use the efficient matrix multiplication implemented in PyTorch \cite{paszke2019pytorch}.

\begin{algorithm}[tb]
    \caption{\textsc{Concept Matrix Search}}
    \label{Alg:CMS}
    \begin{algorithmic}[1]
    \STATE {\bfseries Input:} Batch of image embeddings $\mathrm{I_{|\mathrm{B}|}}$,  labels, all classes $\mathrm{C}$ and concepts $\mathrm{D}$ embeddings.
    \STATE Build $\mathcal{V} \in \mathbb{R}^{|\mathrm{B}|\times |\mathrm{D}|}$, $\mathcal{T} \in \mathbb{R}^{|\mathrm{C}|\times |\mathrm{D}|}$ matrices, store $\mathcal{T}$.
    \FOR{$\mathrm{k}=0,1,2,\dots,|\mathrm{B}|-1$}
    \FOR{$\mathrm{m}=0,1,2,\dots,|\mathrm{C}|-1$} 
    \STATE Compute and store $\cos(\mathcal{V}^\top_{\mathrm{k},\cdot}, \mathcal{T}^\top_{\mathrm{m},\cdot})$
    \ENDFOR
    \STATE Find $\mathrm{m}_{\mathrm{max}}=\underset{\mathrm{m}}{\operatorname{max}}\cos(\mathcal{V}^\top_{\mathrm{k},\cdot}, \mathcal{T}^\top_{\mathrm{m},\cdot})$
    \IF{$\mathrm{label}(\mathrm{k}) = \mathrm{m}_{\mathrm{max}}$}
    \STATE the hypothesis has been proven, increase Accuracy
    \ELSE
    \STATE the hypothesis has been disproved
    \ENDIF
    \ENDFOR
    \STATE \textbf{return} Mean accuracy
\end{algorithmic}
\end{algorithm}

Provided algorithm ensures an increase in the interpretability of the original CLIP model without requiring any additional training. A simple demonstration of our method can be seen in \cref{fig:cms_example}. In this work, we also discover a concept set influence on CMS accuracy, which is presented in Figure \ref{fig:cms_acc}, and discussed in experiments (see Section \ref{sec:cmsexp}). We also present an empirical evidence of the CMS \textit{hypothesis} by considering the classification accuracy of each class separately and show the clear dependence on the concepts set \footnote{\href{https://github.com/icml24/SparseCBM/blob/rebuttal/additional_evaluations/cms_empirical_evidence.ipynb}{repository link}}.
\subsection{Our framework for building CBMs}\label{sec:framework}
\begin{figure*}[t]
\vskip -0.1in
\begin{center}
\centerline{\includegraphics[width=1.5\columnwidth]{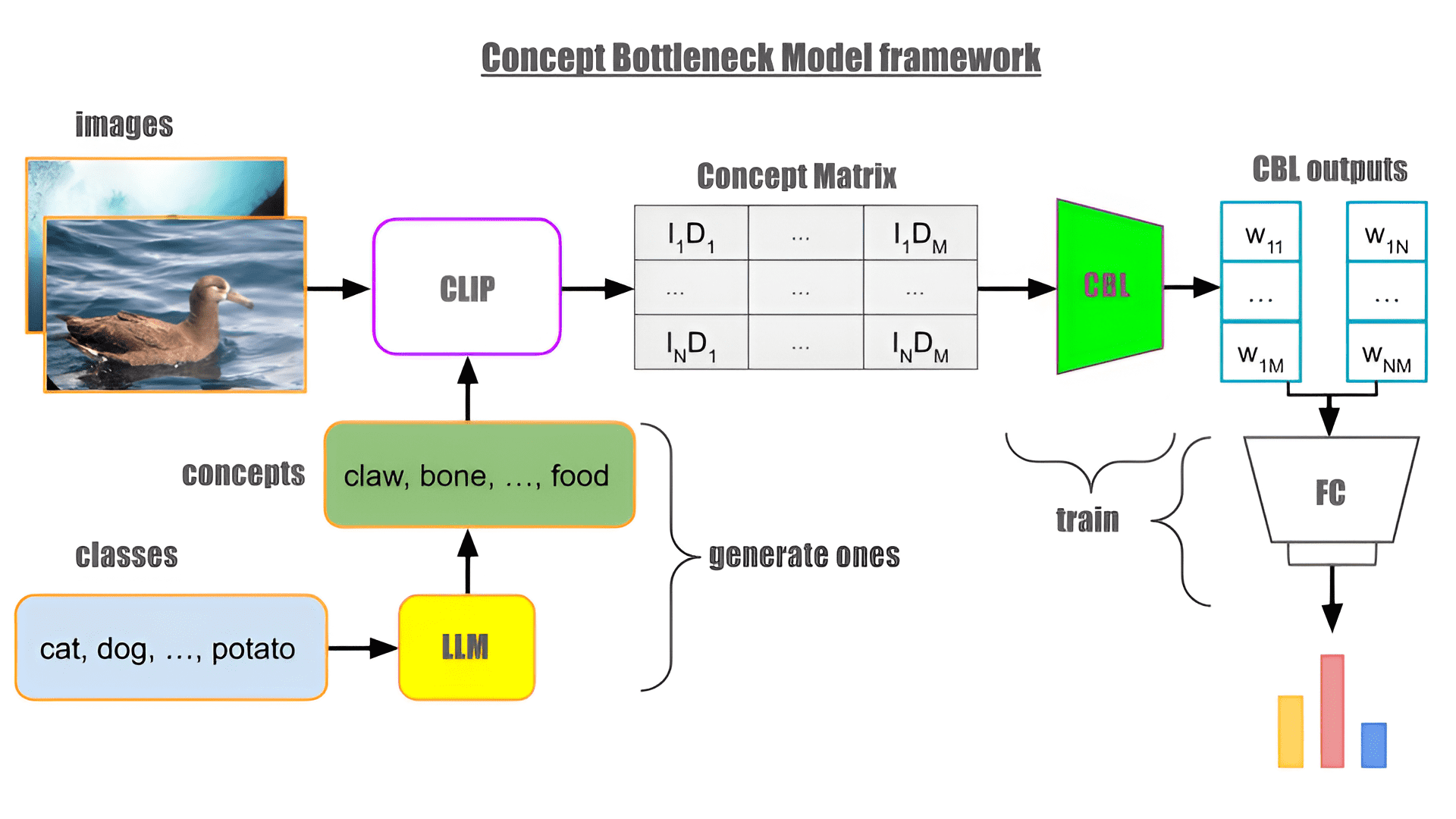}}
\vskip -0.25in
\caption{Overview of our CBM framework.}
\label{fig:framework}
\end{center}
\vskip -0.4in
\end{figure*}
In this section, we propose a framework that creates
Concept Bottleneck Model from pre-trained multi-modal encoder in efficient way with minimum adjusted layers and automated concepts generation for every fine-tuning dataset considered. At first, our CBM builds a model appropriate for image classification, while as a backbone we use bimodal encoder or CLIP-like models that are able to output embeddings or ready-to-use dot-product scores of these embeddings. This means that the final layer of the network is modified based on the number of classes in the dataset. Secondarily, number of concepts and their meaning depend only on labels of dataset and do not changes during further training process. We can summarize building blocks of our framework in the following steps:
\vspace{-0.3cm}
\begin{itemize}
\item \textbf{Step 1:} Pick a proper multi-modal encoder based model to be a backbone (we mainly use OpenAI CLIP-ViT-L/14\footnote{In experiments we use Hugging Face \cite{wolf2020huggingfaces} model present at \href{https://huggingface.co/openai/clip-vit-large-patch14}{https://huggingface.co/openai/clip-vit-large-patch14}.}). Then add two linear layers to it, we refer to the first of them as Concept Bottleneck Layer (CBL) as in \cite{oikarinen2023labelfree}, and the second one is the last FC layer.
\item \textbf{Step 2:} Pick a dataset, create a set of concepts based on class labels and filter\footnote{In experiments we use 'all-mpnet-base-v2' sentence transformer \cite{reimers-2019-sentence-bert} \nocite{reimers-2020-multilingual-sentence-bert} present at \href{https://huggingface.co/sentence-transformers/all-mpnet-base-v2}{https://huggingface.co/sentence-transformers/all-mpnet-base-v2}.} undesired concepts.  
\item \textbf{Step 3:} Pick one from our objective functions mentioned in Sections \ref{sec:sparsecbm}, \ref{sec:ellcbm}, \ref{sec:contrcbm} and two suitable optimizers for obtained architecture (we use Adam \cite{kingma2017adam} and AdamW \cite{loshchilov2019decoupled}).
\item \textbf{Step 4:} Learn CBL with picked at the previous step objective and FC with the Cross-Entropy loss.
\end{itemize}
\vspace{-0.3cm}
More specifically. At \textbf{Step 1}, we consider a CLIP-like model and add two linear layers to it. For a given batch of input images with \textbf{all} created concepts, this model outputs dot-product scores between a particular image and all concepts, i.e., in terms of the CMS algorithm (Section \ref{sec:cms}), we get a batch size $|\mathrm{B}|$ rows of $\mathcal{V}$ matrix. These scores show how sensitive are images from 1 to $|\mathrm{B}|$ to each concept since more suitable concepts get a higher score from CLIP. The first linear layer, we call it CBL, manages concepts informativity, while the second FC make a final prediction. We keep CBL and FC as matrices of shape $|\mathrm{D}|\times|\mathrm{D}|$ and $|\mathrm{D}|\times|\mathrm{C}|$ respectively. The only conceptual difference between these layers is how they are trained. At \textbf{Step 2}, we generate a set of concepts. Exactly as \cite{oikarinen2023labelfree}, we do not rely on any human experts and create concepts manually. One common way is to create them via prompting LLMs about class labels and process LLM's output. We ask GPT-3 \cite{brown2020language}, Llama-2\footnote{In experiments we use free Llama-2 version available at Hugging Face hub \href{https://huggingface.co/TheBloke/Llama-2-70B-Chat-GPTQ}{here}.} \cite{touvron2023llama} for this purposes. Both models demonstrate a good knowledge of which concepts are more preferable for each class. We ask directly the following:

Prompt 1: "List the most important features for recognizing something as a \{label\}. Write them one by one."

Prompt 2: "List the things most commonly seen around a \{label\}. Write them one by one."

Prompt 3: "Give a generalization for the word \{label\}. Answer with a single sentence."

Also, due to the OpenAI API payment rules and the size of Llama-2 models, we use ConceptNet API \cite{article} to create concepts also in automatic manner but with more ease and for free like in \cite{oikarinen2023labelfree,yuksekgonul2023posthoc}. When using ConceptNet we have no opportunity to prompt, instead we use Sentence Transformer \cite{reimers-2019-sentence-bert} and pick the further concepts with the similar to \cite{oikarinen2023labelfree} work algorithm:

1) To represent concepts as a few words we delete all concepts that are longer then 30 letters.

2) We delete all concepts that has cosine similarity with classes more than 0,85 cutoff.

3) We delete all concepts that has cosine similarity with other concepts more than 0,9 cutoff\footnote{0,85 for "concepts-classes" similarity and 0,9 for "concepts-concepts" similarity are our hyperparameters obtained empirically, we do not intend to change them in further experiments}.

4) We delete the concepts with lower average similarity to other concepts to keep more general concepts.

At \textbf{Step 3}, we simply pick the objective functions, we introduce later, to minimize during training of CBL. We insist on using two different objectives and optimizers for CBL and FC. Prior approaches \cite{yuksekgonul2023posthoc,oikarinen2023labelfree} train CBL according to the \textit{independent bottleneck} schema introduced at \cite{koh2020concept}: which means, they firstly train layers which represents concepts activations, and then train a final classifier. 

At \textbf{Step 4}, we, in turn, offer to apply a \textit{sequential bottleneck} schemes: let us train these layers at the same time but apply FC optimizer to all trainable network parameters except of CBL which trains only with its own optimizer. Having CLIP outputs $\psi(x, t) = \left(\langle i, d_1\rangle,\dots,\langle i, d_{|\mathrm{D}|}\rangle \right)^{\mathrm{\top}} \in \mathbb{R}^{|\mathrm{D}|}$ with image embedding $i$ and concept embedding $d_{\mathrm{j}}$ from a dataset $\mathcal{D} = (x, t, l)$, where $x$ stands for visual modality, $t$ stands for \textbf{all} concept words, and  $l$ – class name textual description; loss function for CBL $\mathcal{L}_{\mathrm{CBL}}$; CBL itself $W_{\mathrm{CBL}}\in\mathbb{R}^{|\mathrm{D}|\times|\mathrm{D}|}$; and final layer's Cross-Entropy loss and weights $\mathcal{L}_{\mathrm{CE}}$, $W_{\mathrm{F}}\in\mathbb{R}^{|\mathrm{C}|\times |\mathrm{D}|}$. We can formulate the training period for CBL as follows:
\[
\min \limits_{W_{\mathrm{CBL}}}\mathop{\mathbb{E}}_{(x, t, l)\sim\mathcal{D}}\big[\mathcal{L}_{\mathrm{CBL}}(W_{\mathrm{CBL}}\psi(x, t))\big]_,
\] and, at the previous backpropagation step, we learn FC layer with
\[
\min \limits_{W_{\mathrm{F}}} \mathop{\mathbb{E}}\limits_{(x, t, l)\sim\mathcal{D}}\big[\mathcal{L}_{\mathrm{CE}}(W_{\mathrm{F}}W_{\mathrm{CBL}}\psi(x, t), l) \big]_.
\]
Thus, we train both CBL and FC at the same time but gradient updates from Cross-Entropy loss do not affect $W_{\mathrm{CBL}}$ which comes in handy during the training of bottleneck layers with $\ell_1$ or the Gumbel-Softmax loss functions to make these layers sparse. 
Here we show that $\mathcal{L}_{\mathrm{CE}}$ expects two arguments as a typical supervised loss, while $\mathcal{L}_{\mathrm{CBL}}$ requires only one argument as self-supervised loss function. For both equations we use Adam or AdamW optimizers connected to corresponding layers. The key advantage that our framework offers is how to train this CBM with two optimizers: one for each of presented above task; and which loss functions to choose (see Sections \ref{sec:contrcbm}, \ref{sec:sparsecbm}). And for $\ell_1$-loss training case we consider a regularization term introduced in Section \ref{sec:ellcbm}.

We provide a scheme of our general framework as an architecture with adjusted layers and concept generation in \cref{fig:framework}.

\subsection{Contrastive-CBM}\label{sec:contrcbm}

In this section, we propose a simple adaptation of CLIP loss for training Concept Bottleneck Layers. Using contrastive objective instead of trying to predict the exact words associated with images is a thing that make CLIP-based model popular for zero-shot image classification and image-text similarity. When we create CBM from a pre-trained multi-modal encoder and supply it with the necessary set of concepts, we can also freely train its bottleneck layers with the same contrastive loss \cite{zhai2023sigmoid}. 
But in our setup we force CBL to learn by minimizing not the CLIP "image-text" similarity, but bottleneck layer's outputs $W_{\mathrm{CBL}}\psi(x, t)$ w.r.t. $W_{\mathrm{CBL}}$ parameters. Let $|\mathrm{B}|$ a batch size where given a mini-batch of embeddings $\mathrm{B} = \left((i_1, d_1, c_1), \dots, (i_{\mathrm{|B|}}, d_{\mathrm{|B|}}, c_{\mathrm{|B|}})\right)$. For simplicity, we present the formula in case when number of concepts equals to batch size $|\mathrm{D}| = |\mathrm{B}|$. We also denote $\varphi_{\mathrm{k}} \triangleq \left(\langle i_{\mathrm{k}}, d_{\mathrm{1}}\rangle, \dots, \langle i_{\mathrm{k}}, d_{|\mathrm{B}|}\rangle \right)^\top$ as a vector of dot-product scores between \textbf{all} concepts ($|\mathrm{D}| = |\mathrm{B}|$) and $k$-th image from batch $\mathrm{B}$. As $w_{\mathrm{k}}$ we define a $k$-th row of $W_\mathrm{CBL}$ matrix. Then, our contrastive loss for learning the bottleneck layer can be rewritten as follows:
\begin{align*}\label{eq:contr}
-\frac{1}{2|\mathrm{B}|} \sum_{\mathrm{k}=1}^{|\mathrm{B}|}\Bigg(\log \frac{e^{\alpha \langle w_{\mathrm{k}}, \varphi_{\mathrm{k}}\rangle}}{\sum_{\mathrm{j}=1}^{|\mathrm{B}|}e^{\alpha \langle w_{\mathrm{k}}, \varphi_{\mathrm{j}}\rangle}} + \log \frac{e^{\alpha \langle w_{\mathrm{k}}, \varphi_{\mathrm{k}}\rangle}}{\sum_{\mathrm{j}=1}^{|\mathrm{B}|}e^{\alpha \langle w_{\mathrm{j}}, \varphi_{\mathrm{k}}\rangle}} \Bigg)_.
\end{align*}
The scalar $\alpha$ is parameterized as $\exp{\tilde{\alpha}}$\footnote{In OpenAI CLIP $\tilde{\alpha} = 1.155$ which we use for CMS backbone CLIP, while the value of $2.659$ is used for CBM}. We define a "Contrastive-CBM" as a special case of our framework's Concept Bottleneck Model with contrastive objective $\mathcal{L}_{\mathrm{CBL}}$ for bottleneck layers of presented above form. We also note that this kind of loss can be effectively implemented for distributed learning  \cite{chen2023discoclip,zhai2023sigmoid}.

\subsection{Sparse-CBM}\label{sec:sparsecbm}

In this section, we propose the main objective function for training our models. 
Firstly, we define a Gumbel-Softmax distribution \cite{jang2017categorical,DBLP:journals/corr/MaddisonMT16} as follows: let $z$ be a categorical variable with probabilities $\pi_1,\cdots, \pi_{|\mathrm{D}|}$, i.e., $|\mathrm{D}|$-dimensional one-hot vector from probability simplex $\Delta^{|\mathrm{D}|-1} \triangleq \{ \pi \in \mathbb{R}_+^{|\mathrm{D}|} : \sum_{i=1}^{|\mathrm{D}|} \pi_i =1 \}$. Then, Gumbel-Max trick \cite{Gumbel1954StatisticalTO,maddison2015a} allows us to sample $z$ from a categorical distribution with class probabilities $\pi = (\pi_1, \cdots, \pi_{|\mathrm{D}|})$:
\vskip -0.15in
\[
z = \mathbf{1}\left(\arg\max \limits_{\mathrm{k}}\big[g_{\mathrm{k}} + \log \pi_{\mathrm{k}}\big] \right)_,
\]
 \vskip -0.15in
where $g_1,\dots, g_{|\mathrm{D}|}$ are i.i.d. samples drawn from $\mathrm{Gumbel}(0,1)$, which, in turn, can be obtained from $u \in \mathrm{Uniform}(0,1)$ by sampling $g = -\log \log u$, (we denote indicator $\mathbf{1}$ as one-hot function). After applying a softmax function as a continuous, differentiable approximation to $\arg \max$ we obtain a Gumbel-Softmax distribution, a continuous distribution over the simplex that can approximate samples from a categorical distribution. Then we intend to build a contrastive loss for CBL outputs as follows: 
\begin{align*}\label{eq:gumbellos}
-\frac{1}{2|\mathrm{B}|} \sum_{\mathrm{k}=1}^{|\mathrm{B}|}\Bigg(&\log \frac{e^{\left(\log(\alpha \langle w_{\mathrm{k}}, \varphi_{\mathrm{k}}\rangle) + g_{\mathrm{k}}\right)/\tau}}{\sum_{\mathrm{j}=1}^{|\mathrm{B}|}e^{\left(\log(\alpha \langle w_{\mathrm{k}}, \varphi_{\mathrm{j}}\rangle) + g_{\mathrm{j}}\right)/\tau}} \notag \\ 
&+ \log \frac{e^{\left(\log(\alpha \langle w_{\mathrm{k}}, \varphi_{\mathrm{k}}\rangle) + g_{\mathrm{k}}\right)/\tau}}{\sum_{\mathrm{j}=1}^{|\mathrm{B}|}e^{\left(\log(\alpha \langle w_{\mathrm{j}}, \varphi_{\mathrm{k}}\rangle) + g_{\mathrm{j}}\right)/\tau}} \Bigg)_.
\end{align*}

We use the provided loss to represent a Concept Bottleneck Layer's logits as a categorical variable, i.e., sparse vectors which are highly interpreted \cite{pmlr-v139-wong21b}. With Gumbel-Softmax structure $\mathcal{L}_{\mathrm{CBL}}$ achieves a sparse learning of bottleneck layers which increases an interpretability of them: key ingredient to provide a sparsification of CBL output vectors is temperature $\tau$. Presented distribution is smooth for $\tau > 0$, it has a defined gradients w.r.t. $w_{\mathrm{k}}$ parameters. By replacing a categorical samples with Gumbel-Softmax samples we can backpropagate through Concept Bottleneck layers. For low temperatures ($\tau < 0.5$) the expected value of Gumbel-Softmax distribution approaches the expected value of a categorical random variable with the same logits, i.e., becomes a one-hot. As the temperature increases, the expected value converges to a uniform distribution over the categories. In practice, we start at a high temperature and anneal to a small but non-zero temperature.
\begin{figure}[h]
\begin{center}
\centerline{
\includegraphics[width=0.6\columnwidth]{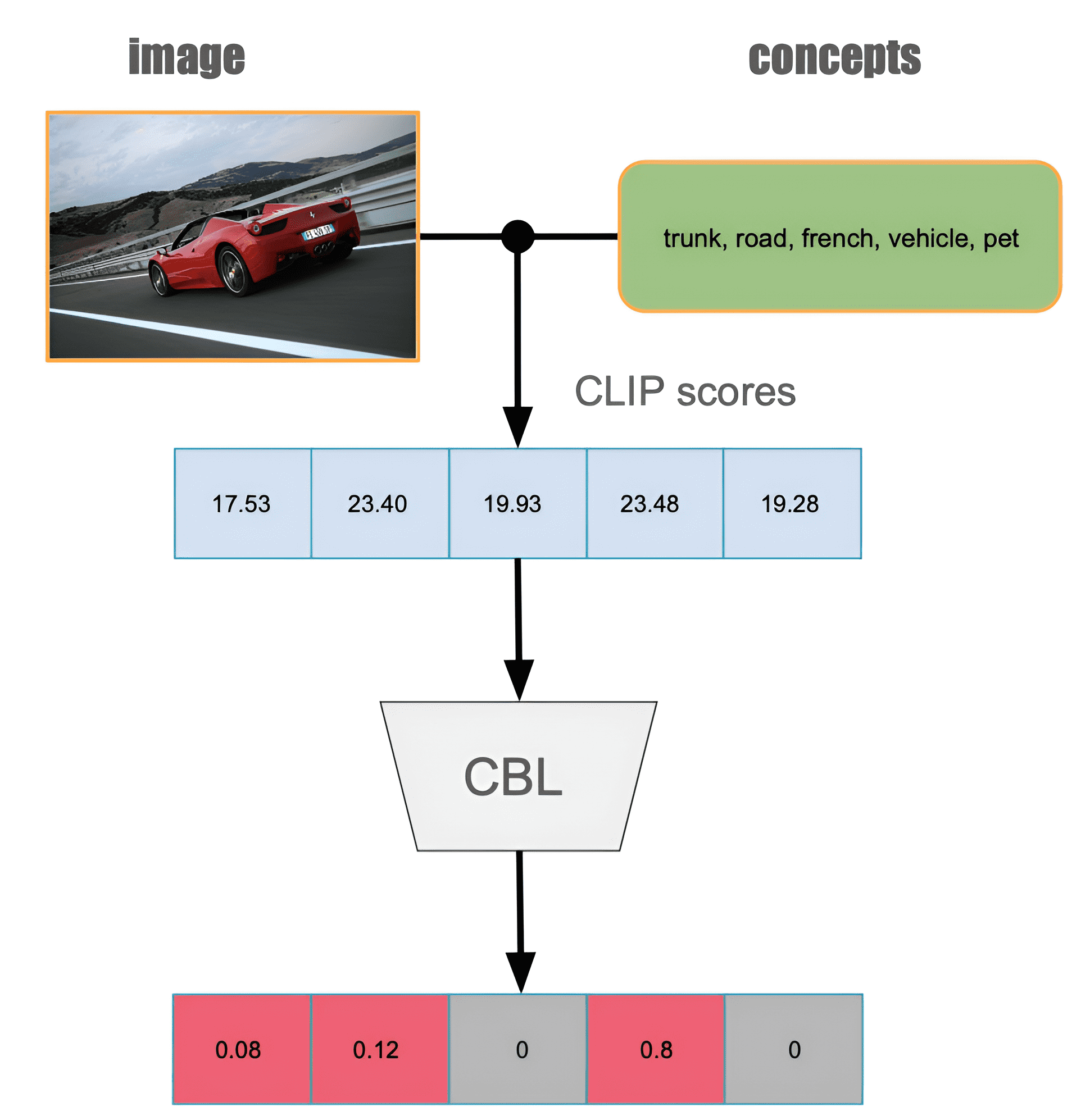}}
\caption{Visualization of Sparse-CBM Concept Bottleneck Layers.}
\label{fig:gumbel_example}
\end{center}
\vskip -0.35in
\end{figure}
If $\tau$ is a learned parameter (rather than annealed via a fixed schedule), this scheme can be interpreted as entropy regularization \cite{szegedy2015rethinking,pereyra2017regularizing}. A high-level explanation and derivation of Gumbel activations can bee seen at \cite{alexandridis2022longtailed}.

\subsection{$\boldsymbol\ell_1$-CBM}\label{sec:ellcbm}

We also train Concept Bottleneck with $\ell_1$-loss function. Before \cite{yuksekgonul2023posthoc} proposed to use elastic-net penalty and \cite{oikarinen2023labelfree} to use a weighted mean between Frobenius norm and element wise matrix norm. Instead, we show a promising result only with $\ell_1$ penalizing. $\ell_1$ penalty shows an ability to sparsify a trainable layers: the optimizer tries to minimize the loss by reducing the weights that have a non-zero gradient, this leads to some weights being set to exactly zero, which effectively removes the corresponding features from the output of CBL.

We begin with defining the following optimization problem:
\[
\min \limits_{W_{\mathrm{CBL}}} \mathop{\mathbb{E}}\limits_{(x, t, l)\sim\mathcal{D}}\big[\mathcal{L}_{\mathrm{CE}}(W_{\mathrm{F}}W_{\mathrm{CBL}}\psi(x, t), l) + \frac{\lambda}{|\mathrm{D}|}\Omega(W_{\mathrm{CBL}}) \big]_,
\]\label{eq:l1probl}where $\Omega(W_{\mathrm{CBL}})$ stands for regularization term. We use:
\[
\Omega(W_{\mathrm{CBL}}) = \|W_{\mathrm{CBL}}\|_1
\] 
with the only one parameterization $\lambda$. 

\section{Experiments}\label{sec:exps}

\paragraph{Baseline.} Here we provide a basic information about models and algorithms with which we compare ours.

1) We evaluate the effectiveness of described framework (Section \ref{sec:framework}), in particular, Sparse-, Contrastive-, $\ell_1$-CBMs by comparing it with prior Label-free \cite{oikarinen2023labelfree}, Post-hoc CBM \cite{yuksekgonul2023posthoc}, LaBo \cite{yang2023language} frameworks on the same datasets \cite{5206848,wah_branson_welinder_perona_belongie_2011,krizhevsky2009learning,7968387} but with different backbone models: while in Label-free framework experiments conducted with ResNet50 \cite{he2015deep} and CLIP(ResNet50) \cite{radford2021learning}, we use only CLIP architecture variants as the backbone. We notice, that both frameworks obtain a quite similar concepts because Label-free also include an option to generate concepts using ConceptNet API. Also Label-free model is trained by two algorithms in sequential way: firstly, they train a CBLs with cos-cubed similarity introduced at \cite{oikarinen2023labelfree}, then the last FC layer is trained with GLM-SAGA \cite{pmlr-v139-wong21b} solver. We also present a comparison with Linear Probing, when the one linear layer after the CLIP is trained to perform classification. The detailed explanation of the experiment with probing we show in \cref{sec:appendix}.

2) We evaluate the effectiveness of described Concept Matrix Search algorithm (Section \ref{sec:cms}) by comparing it with prior Visual Classification via Description work which we refer to as "DescriptionCLS" method in \cref{tab:cbms_tab}. Both methods are similar in case they are able to discriminate between classes by selecting the one with the highest score provided by some \textit{rule} and both of them do not train any model and leverage only an empirical formulae to predict targets in more interpreted way. Our \textit{rule} is the CMS (see \cref{Alg:CMS}), while "DescriptionCLS" method relies on the function which gives a score to the class based on how many concepts of this class are sensitive to the image of this very class. Mentioned sensitivity is measured as a log probability that concept pertains to the image in accordance with cosine similarity. CMS also leverages this measure, but it does not create a direct class-concept mapping, instead we make a classification in more general way given only a batch of images and all concepts with classes. Besides, Concept Matrix Search plays a role of more general classifier and we compare them below which is informative since both methods use the same CLIP model as a backbone. Furthermore, it is essential to provide a comparison with zero-shot image classification. In this setting, we treat a batch of images and class names as input to be encoded by CLIP, which, in turn, outputs the Image-Class Matrix. By performing an argmax operation we determine the most related class.

\setcounter{table}{1}
\begin{table*}[t]
\caption{Comparison of Bottleneck model's performance on main classification datasets. We see an improvement of our Sparse-CBM over other architectures on CIFAR10, CIFAR100 and CUB200 datasets.}
\label{tab:cbms_tab}
\begin{center}
\vskip -0.15in
\begin{small}
\begin{sc}
\begin{tabular}{lccccr}
\toprule
Model & CIFAR10 & CIFAR100 & ImageNet & CUB200 & Places365 \\
\midrule
Sparse-CBM (Ours)    & \textbf{91.17\%} & \textbf{74.88\%} & 71.61\% & \textbf{80.02\%} & 41.34\% \\
$\ell_1$-CBM (Ours) & 85.11\% & 73.24\% & 71.02\% & 74.91\% & 40.87\%\\
Contrastive-CBM (Ours)   & 84.75\% & 68.46\% & 70.22\% & 67.04\% & 40.22\% \\
Label-free CBM    & 86.40\% & 65.13\% & \textbf{71.95\%} & 74.31\% & \textbf{43.68\%}    \\
Post-hoc CBM (CLIP)   &  83.34\%       &   57.20\%      &     62.57\%         &   63.92\%      &       39.66\%        \\
LaBo (full-supervised)             & 87.90\%&       69.10\%    &    70.40\%     & 71.80\% & 39.43\%
\\
\hline
Linear Probing             & 96.12\%&       80.03\%     &    83.90\%     & 79.29\% & 48.33\%     \\
\bottomrule
\end{tabular}
\end{sc}
\end{small}
\end{center}
\vskip -0.1in
\end{table*}
\begin{table*}[t]
\caption{Comparison of CMS and "DescriptionCLS" performance on main datasets.}
\label{tab:cms_tab}
\begin{center}
\vskip -0.15in
\begin{small}
\begin{sc}
\begin{tabular}{lccccr}
\toprule
Method & CIFAR10 & CIFAR100 & ImageNet & CUB200 & Places365 \\
\midrule
Concept Matrix Search (Ours)    & \textbf{85.03\%} & 62.95\% & \textbf{77.82\%} & \textbf{65.17\%} & 39.43\% \\
DescriptionCLS  &    81.61\%     &      \textbf{68.32\%}    &     75.00\%     &   63.46\%    &   40.55\%    \\
Zero-shot  &    81.79\%     &      52.84\%    &     76.20\%     &   62.63\%    &   \textbf{41.12\%}    \\
\bottomrule
\end{tabular}
\end{sc}
\end{small}
\end{center}
\vskip -0.3in
\end{table*}

\subsection{CBM experiments}\label{sec:cbmexp}

\paragraph{Set of concepts.} We have generated a large sets: one with 6,500 concepts using Llama-2 and the second with 5,000 concepts using ConceptNet \cite{article} API and processed them with 'all-mpnet-base-v2' sentence transformer to find similarity cutt-offs mentioned at \textbf{Step 2} (see \cref{sec:framework}). For the biggest set we prepared all class labels to create a concepts from them. Moreover, for each dataset, a corresponding set of concepts also provided in \cref{tab:concepts_size}. We call the largest set of concepts "All Concepts" which can be seen in the \cref{fig:cms_acc}. Each dataset has its own corresponding set of concepts generated from dataset classes according to \textbf{Step 2}. For fair comparison, provided in \cref{tab:cbms_tab} models were trained \textit{only} with ConceptNet generated concepts.

\vskip 0.1in
\paragraph{Models.} CLIP-ViT-L/14 and B/32\footnote{Mentioned CLIP-ViT-B/32 model can be found at Hugging Face hub \href{https://huggingface.co/openai/clip-vit-base-patch32}{https://huggingface.co/openai/clip-vit-base-patch32}} as a backbone for all CBM experiments. Depending on the dataset, the size of these models varies. The smallest one B/32 configuration
\begin{wrapfigure}{l}{0.285\textwidth}
\vskip -0.1in
\setcounter{table}{0}
\captionof{table}{Dependence of concept set size on the dataset.}
\label{tab:concepts_size}
 \vskip 0.15in
\begin{center}
\vskip -0.2in
\begin{small}
\begin{sc}
\begin{tabular}{lr}
\toprule
 Dataset & \# of concepts   \\
\midrule
CIFAR10  & 120   \\
CIFAR100 & 944 \\
ImageNet & 4,751 \\
CUB200 & 926 \\
Places365 & 2,900\\
"All concepts" & 5,051 \\
\bottomrule
\end{tabular}
\end{sc}
\end{small}
\end{center}
\vskip -0.2in
\end{wrapfigure}
 for CIFAR10 has a size of 151,3 million parameters and 50,000 of trainable parameters, which is equivalent to 0,6GB. The biggest one is L/14 for ImageNet-1K: 455 million parameters; 27,3 million of trainable parameters and size of 1,81GB. For clarity, we show a precise information in \cref{tab:backbone_nets} (before the training). And in \cref{tab:cbms_tab} we report the results for CLIP-ViT-L/14 configuration.

\vskip 0.1in
\paragraph{Hardware and Schedule.}
We trained our models on one machine with 4 NVIDIA A100-SXM4-80GB and A100-PCIE-40GB GPUs, two of each type. For our CBM configurations each training step with default learning rates, and mentioned in Tables \ref{tab:backbone_nets}, \ref{tab:concepts_size} data, took less than 1 second for CLIP-ViT-B/32 backbone configuration and slightly longer for CLIP-ViT-L/14. We trained each of Sparse, $\ell_1$, Contrastive-CBMs on ImageNet-1K \cite{russakovsky2015imagenet} for 20,000 steps which takes a bit more than 5,5 hours for each model (see \cref{fig:curves}). Our implementations do not depend on specific hardware architectures.

\begin{figure}[t]
\vskip 0.05in
\begin{center}
\centerline{\includegraphics[width=\columnwidth]{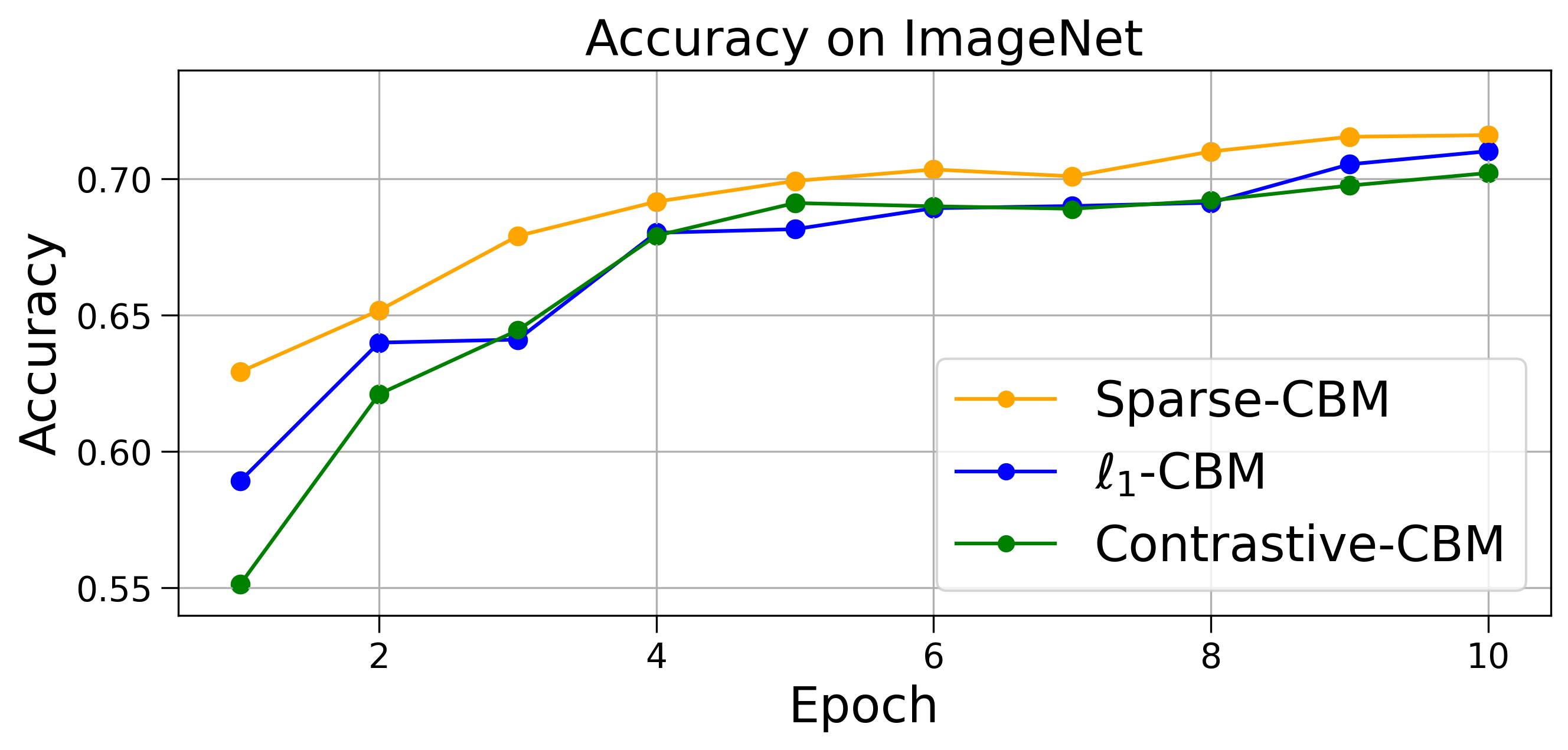}}
\vskip -0.15in
\caption{Comparison of our CBM methods on the ImageNet validation subset.}
\label{fig:cbm_acc}
\end{center}
\vskip -0.45in
\end{figure}
\vskip 0.1in
\paragraph{Workload.} We support a \textit{sequential bottleneck} scheme from \cite{koh2020concept}. Indeed, at first we make a gradient updates of $W_{\mathrm{F}}$ and $W_{\mathrm{CBL}}$ weights w.r.t. to their parameters independently, and then, during the next forward pass, FC receives as input a logits from the updated CBL. Thus, after an every iteration of CBL and FC minimization problems, a new concept representation is provided for classifier.   

\subsection{CMS experiments}\label{sec:cmsexp}

In this section we compare two algorithms: our CMS \ref{Alg:CMS} and main method proposed in \cite{menon2022visual}. Both use a basic CLIP model capabilities. We test them on the same CLIP-ViT-L/14 configuration. We show a superiority of our method on several datasets which can be seen in \cref{tab:cms_tab}. Authors of \cite{kazmierczak2023clipqda} measured an accuracy dependence when a particular number of concepts was nullified, and \cite{chauhan2023interactive} investigated an impact of several relative concepts. We, instead, have measured an accuracy dependence on different concept sets which are: almost synonims ("concept-class" similarity is 0,95 instead of 0,85), generated by ConceptNet as usual, a one set of many concepts (most of them a useless for CIFAR10 data), and 3,000 of random generated words. The results are present in \cref{fig:cms_acc}.


\section{Discussion}\label{sec:disc}

\cref{tab:cbms_tab} demonstrates the superiority of Sparse-CBM in comparison with Label-free on all datasets except of ImageNet and Places365. Moreover, Contrastive-CBM has the lowest overall score. We interpret these observations as follows:

\begin{figure}[t]
\vskip 0.05in
\begin{center}
\centerline{\includegraphics[width=\columnwidth]{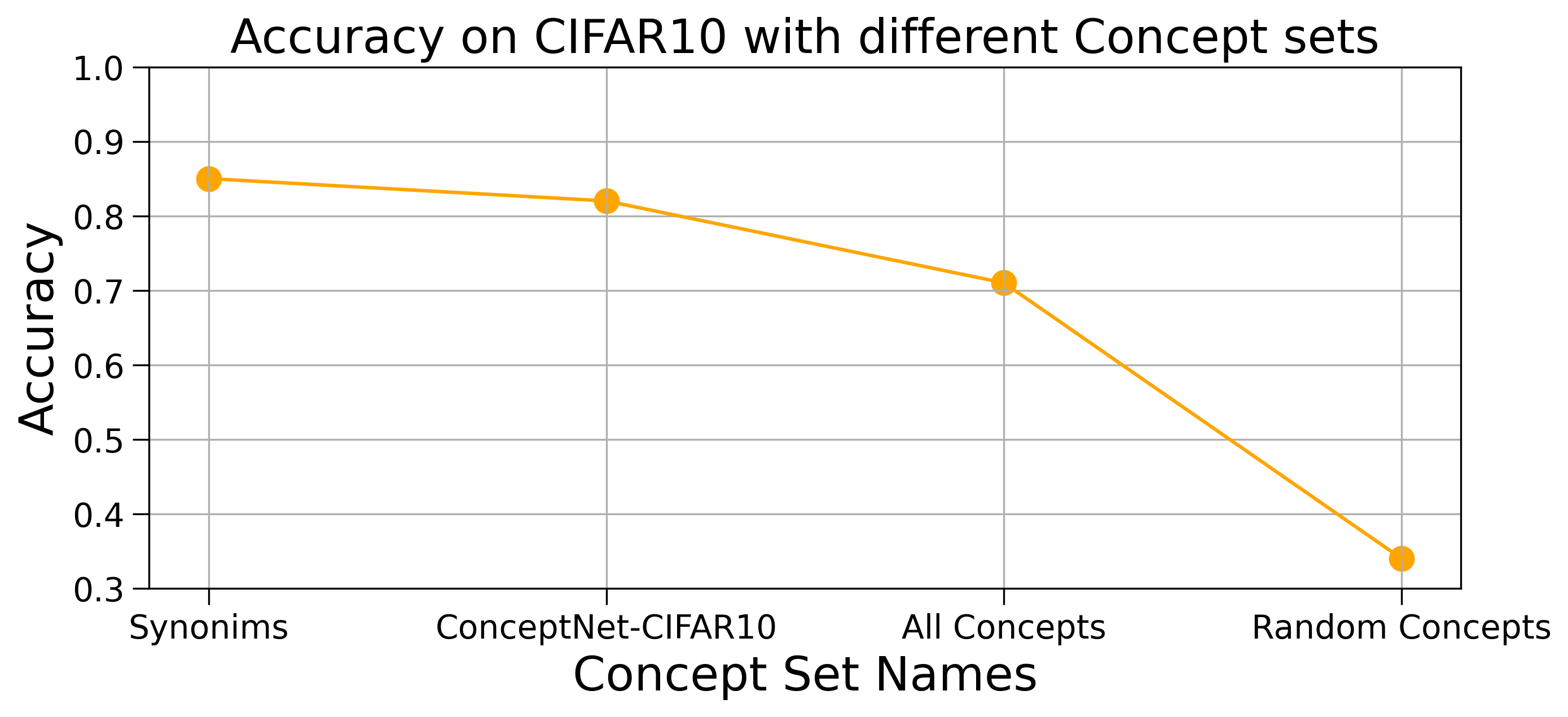}}
\vskip -0.15in
\caption{Dependence of CMS accuracy on a concept set.}
\label{fig:cms_acc}
\end{center}
\vskip -0.5in
\end{figure}

1) Due to \cref{tab:concepts_size}, the proportion between CBL size and FC size differs. While datasets such as CIFAR10, CIFAR100 contain $\approx$ 10 times more concepts than classes, this number, and therefore layers size proportion, is around 4.6–4.7 times larger for ImageNet and Places365. We insist that the sparsity of inner layers brings more benefits with an increase of CBL dimension compared to the output dimension of last fully connected layer.

2) Contrastive-CBM method lacks the interpretability of concept representation, i.e., sparsity of CBL (see a deatailed comparison of its interpretability in \cref{sec:concepts_interpretability}).
We also refer to the CMS algorithm and show a decrease in accuracy depending on the "quality" of concepts, which makes our method, and similar ones, less universal. On the other hand, with ''good'' set of concepts for each dataset we show that CMS outperforms Zero-shot CLIP. Which show us that the model gets additional ''confidence'' when predicting not only the most similar class, but also when obtaining several significant concepts per image.

\subsection{Ablation study}\label{sec:abl}
Components introduced in our framework cause a drop of final performance in comparison to Linear Probing (see \cref{tab:cbms_tab_new}) while ensuring interpretability which we present in \cref{sec:concepts_interpretability}. Our method and its intuition of making inner representation more sparse are clear, and it turns out to be a promising extraction of meaningful concepts. Nevertheless, ''hard'' (i.e., making samples directly one-hot) sampling in contrastive Gumbel-Softmax loss rather harms the performance of Sparse-CBM than makes it better. At the same time, Contrastive-CBM performs a double softmax on CBL which ensures no sparsity and achieves poorer result in accuracy which, again, claims our hypothesis about the utility of sparse inner representations. Both of CBM framework and CMS method rely on the generated set of concepts. With concepts that are better tailored to dataset we show higher results in terms of classification accuracy, on the other hand, we want to keep concepts more diverse which is empirically defined in \cref{sec:framework} and in the prior \cite{oikarinen2023labelfree} work, then, it is not essential for us to change the amount of concepts manually, we would rather keep the generated ones. 

\subsection{Limitations}\label{sec:lims}

In addition to the achieved accuracy performance, we report main limitations of our framework and Algorithm \ref{Alg:CMS}. LLM proposed in the CBM framework still does not affect the classification process, i.e., it does not support concept generation over time and only works at \textbf{Step 2}. Both CMS and CBM variants do not modify the CLIP latent space which makes our approach less versatile. Although we contribute to understanding of sparse CBLs, the effectiveness of bottleneck models with also sparse FC is still uncovered. The same applies to the problem of editing the Concept Bottleck in end-to-end manner (see \cref{sec:bottls}). Finally, CMS method can be useful as an add-on to CLIP, but, in the long run, cannot compete with CBM approaches due to the situation shown in \cref{fig:cms_acc}.   
\nocite{JMLR:v9:vandermaaten08a}

\section*{Impact Statement}
This paper presents work whose goal is to advance the field of Machine Learning. There are many potential societal consequences of our work, none of which we feel must be specifically highlighted here.

\bibliography{ref}
\bibliographystyle{icml2024}

\newpage
\appendix
\onecolumn
\section{Appendix}\label{sec:appendix}
We compare our architectures for Sparse-, $\ell_1$-, Contrastive- CBMs with Post-hoc CBM \cite{yuksekgonul2023posthoc} and LaBo \cite{yang2023language} in terms of accuracy in the downstream image classification task. While the concepts interpretability is compared with the backbone multi-modal backbone model properties in \cref{sec:concepts_interpretability}. Since Post-hoc frameworks also allows to create the Concept Bottleneck Model above the standard backbone, we evaluate this method on CLIP-ViT-L/14 model for fair comparison, at the same time, LaBo architectures are built with the same backbone by defalut. Moreover, we include a results on Linear Probing CLIP-ViT-L/14. In this setting, we add only one linear layer after the Image-Class matrix returned by CLIP. The updated results can be viewed in the \cref{tab:cbms_tab}.


"full-supervised" setting in LaBo means that the architecture is trained on all available images. We note this due to considered by \cite{yang2023language} zero-shot and N-shot learning regimes.

\section{Additional experiments}
In this section we report the additional evaluation results of CBM framework (Section \ref{sec:framework}) and CMS algorithm (Section \ref{sec:cms}).

\subsection{Concept Bottleneck Model training regime}
\begin{figure}[h] 
\centering
   \begin{subfigure}
     \centering
    \includegraphics[width=0.5\linewidth]{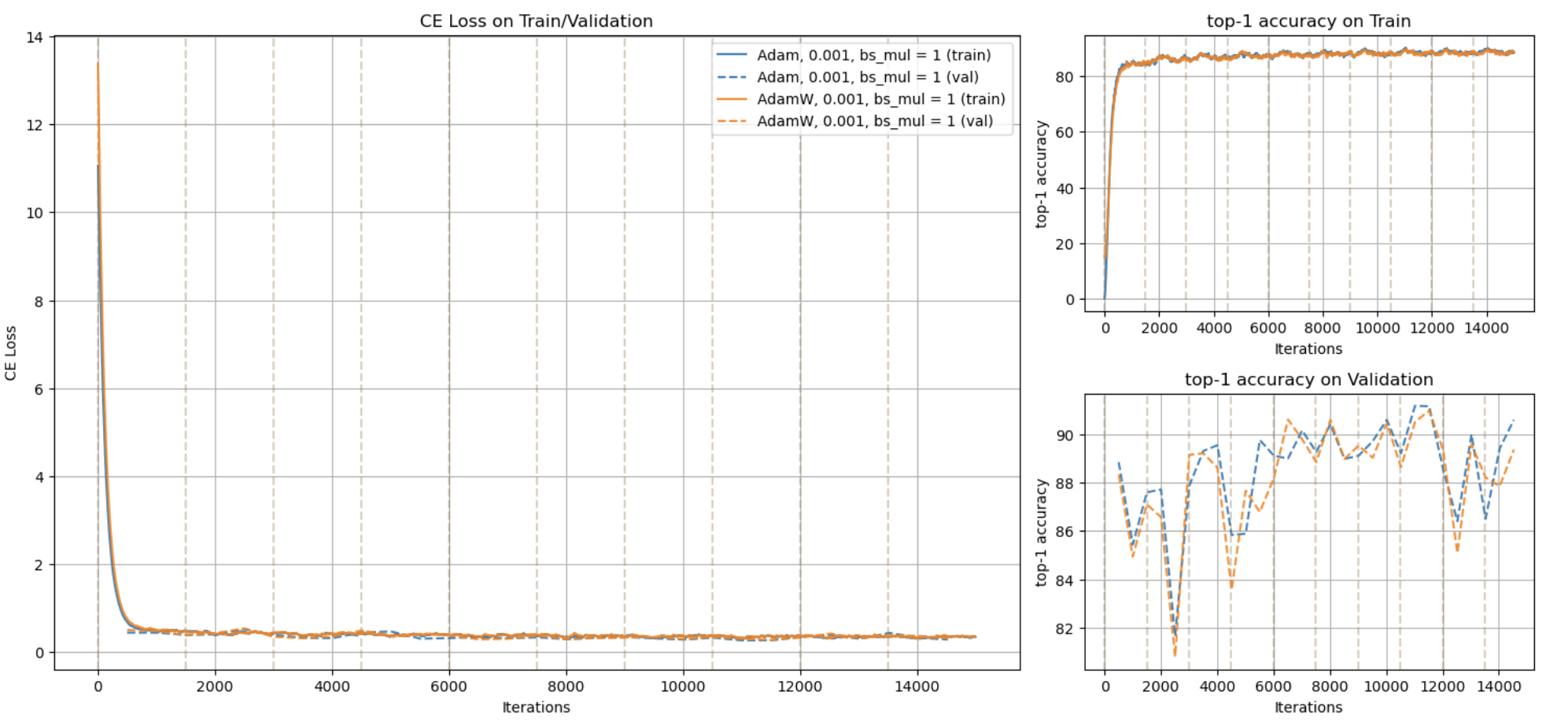}
    \caption{The best Sparse-CBM result on CIFAR10.}
    \label{fig:sparse_best_cifar}
    \end{subfigure}
    \begin{subfigure}
    \centering
      \includegraphics[width=0.5\linewidth]{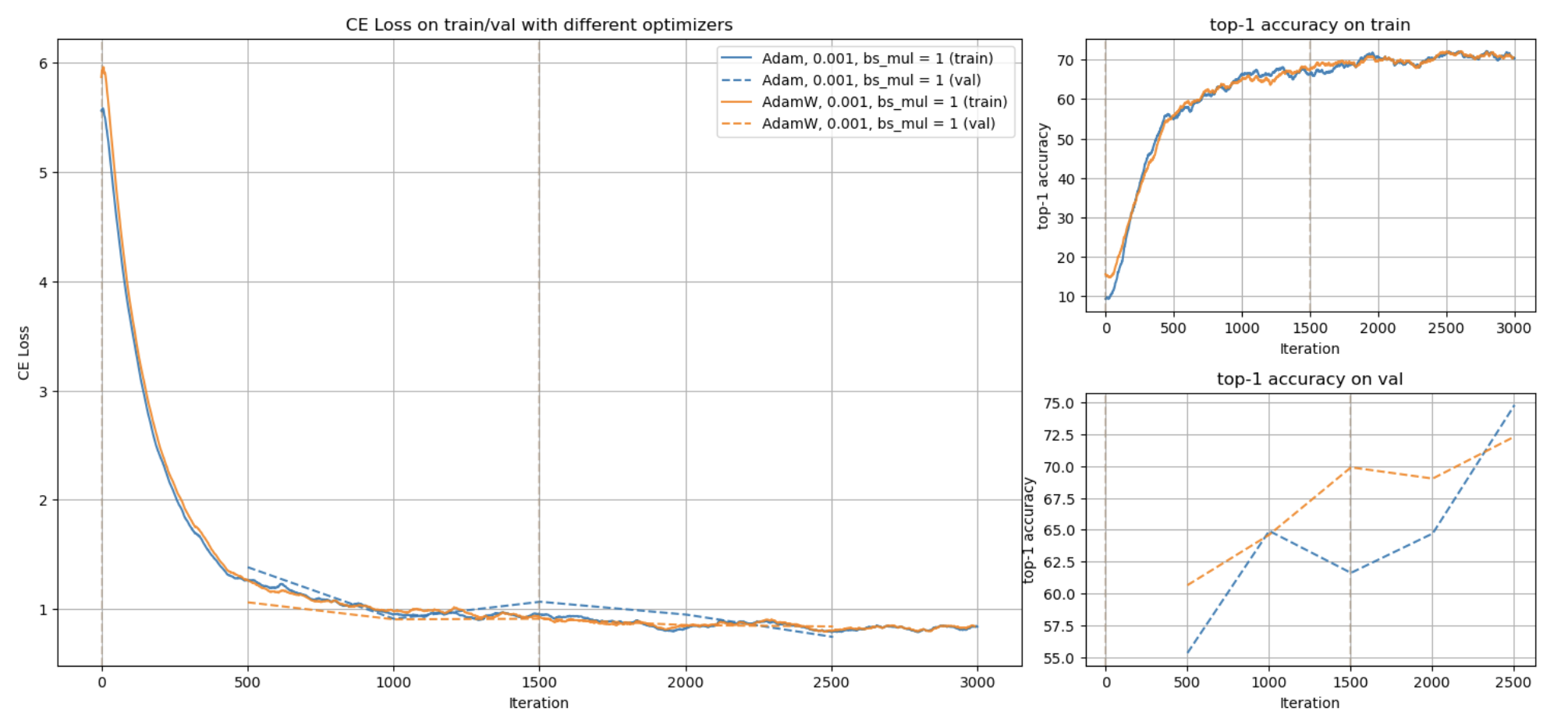}
    \caption{The best Sparse-CBM result on CUB200.}
    \label{fig:sparse_best_imnet}
    \end{subfigure}
    \begin{subfigure}
     \centering
  \includegraphics[width=0.5\linewidth]{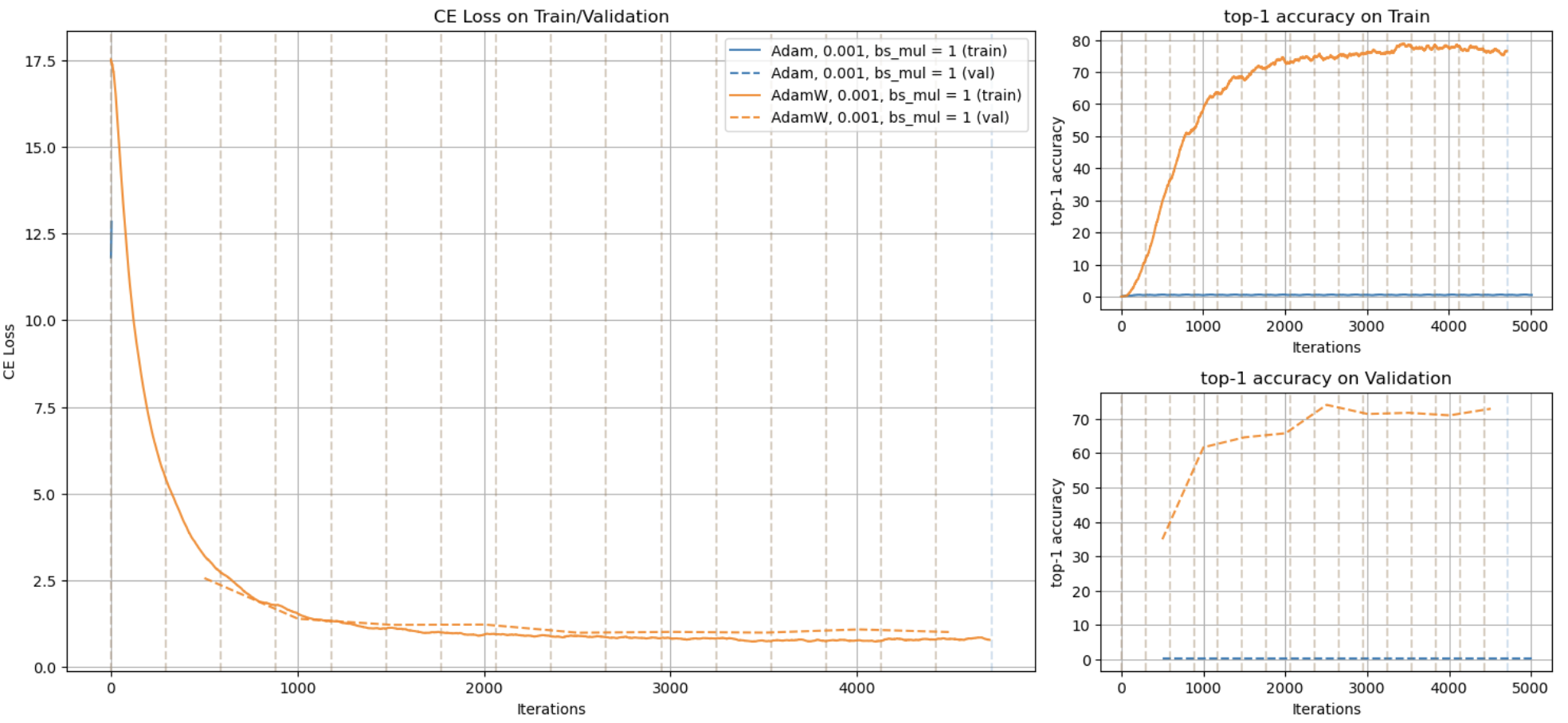}
    \caption{The best Sparse-CBM result on ImageNet-1K.}
    \label{fig:sparse_best_cub}
    \end{subfigure}
    \caption{Overview of Sparse-CBM training regime on several datasets. We observe a similar training loss curves on all data. But for ImageNet, training starts with a smaller learning rate.}\label{fig:curves}
\end{figure}

We provide a several plots (\cref{fig:sparse_best_cifar,fig:sparse_best_cub,fig:sparse_best_imnet}) monitored during Sparse-CBM training on CIFAR10, CUB200 and ImageNet-1K.

\section{Additional data}

\begin{table}[h]
\caption{Backbone models configurations. Intersections indicate the size of the model with the appropriate configuration.}
\label{tab:backbone_nets}
\vskip 0.15in
\begin{center}
\begin{small}
\begin{sc}
\begin{tabular}{lcccr}
\toprule
B/32 & L/14 & Dataset   \\
\midrule
 0.57GB & 1.63GB &  CIFAR10 \\
 0.58GB& 1.64GB & CIFAR100 \\
 0.68GB &  1.74GB & ImageNet \\
 0.58GB& 1.64GB & CUB200 \\
 0.61GB& 1.67GB & Places365 \\
\bottomrule
\end{tabular}
\end{sc}
\end{small}
\end{center}
\vskip -0.1in
\end{table}

\section{Visualizations}
\begin{figure}[ht]
\vskip 0.2in
\begin{center}
\centerline{\includegraphics[width=0.5\columnwidth] {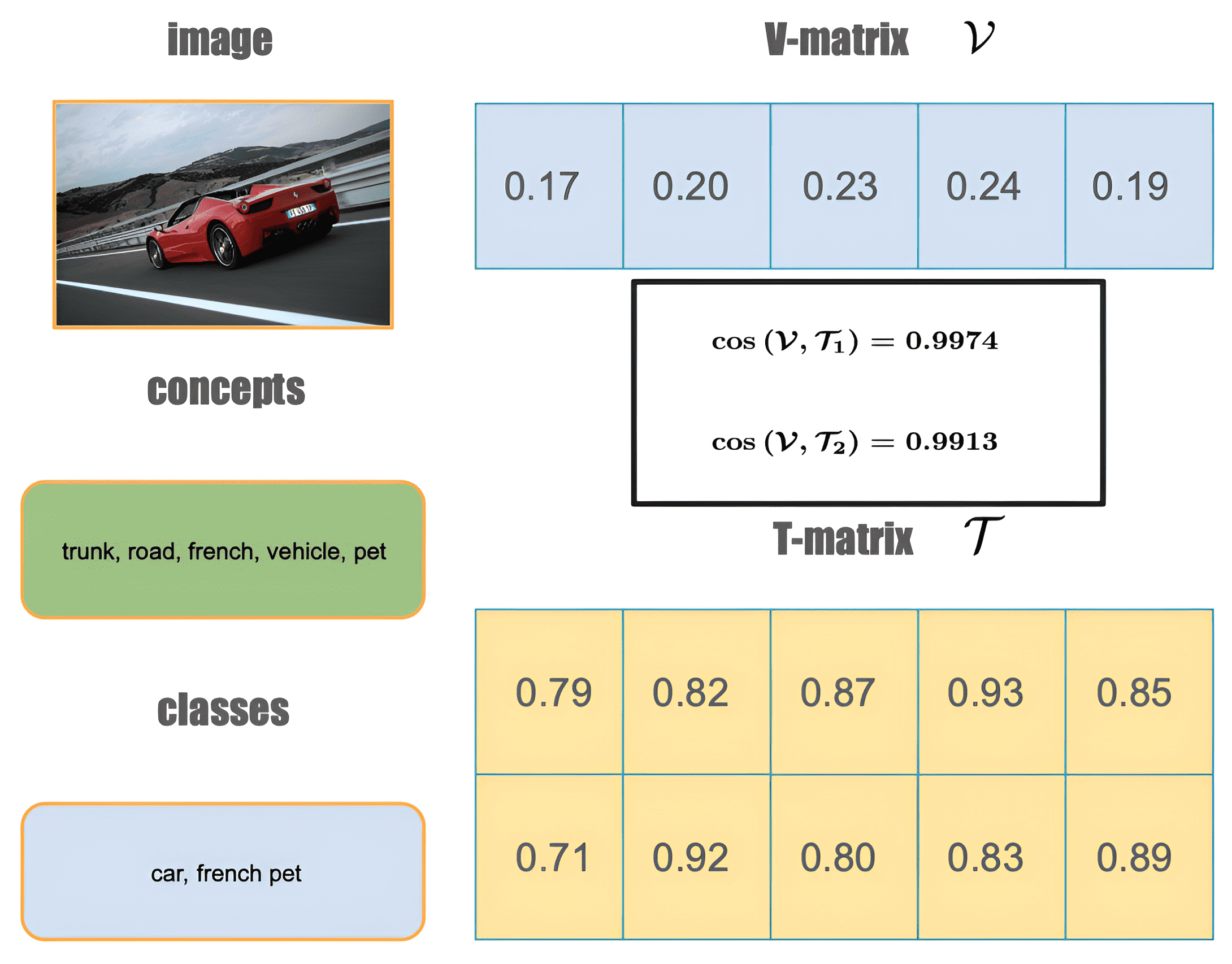}} 
\caption{Visualization of Concept Matrix Search \textit{hypothesis} for a simple case of 1 image, 5 concepts and 2 classes.}
\label{fig:cms_example}
\end{center}
\vskip -0.2in
\end{figure}

\subsection{CLIP latent space analysis}
\begin{figure}[h]
\begin{center}
\centerline{\includegraphics[width=0.5\columnwidth]{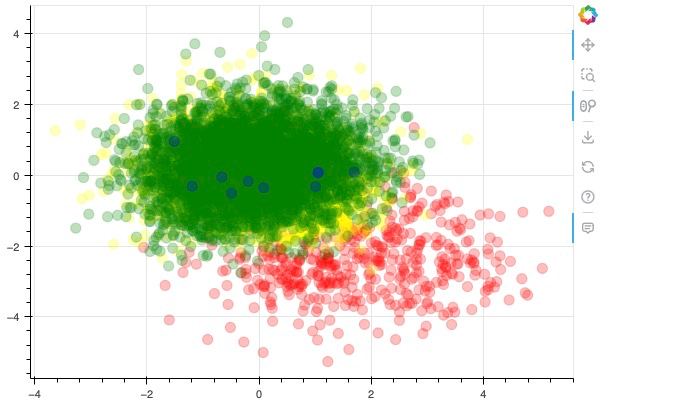}} 
\caption{Visualization of CIFAR10 t-SNE. Green points refer to concepts projection, blue ones – to classes, red – images and yellow are random words.}
\label{fig:tsne}
\end{center}
\end{figure}

In this section, we present more experiments with CLIP latent space. Firstly, using CLIP, we build a two dimensional t-SNE map of CIFAR10 embeddings along with their concepts and classes. An interesting result we show, is that this space not just divided into two clusters: one corresponding to a textual modality and the second to a visual one, but also, if we add a random words projection to t-SNE, we observe its overlap with concepts. Corresponding experiment is present in \cref{fig:tsne}

\begin{figure}[t]
\begin{center}
\centerline{\includegraphics[width=\columnwidth]{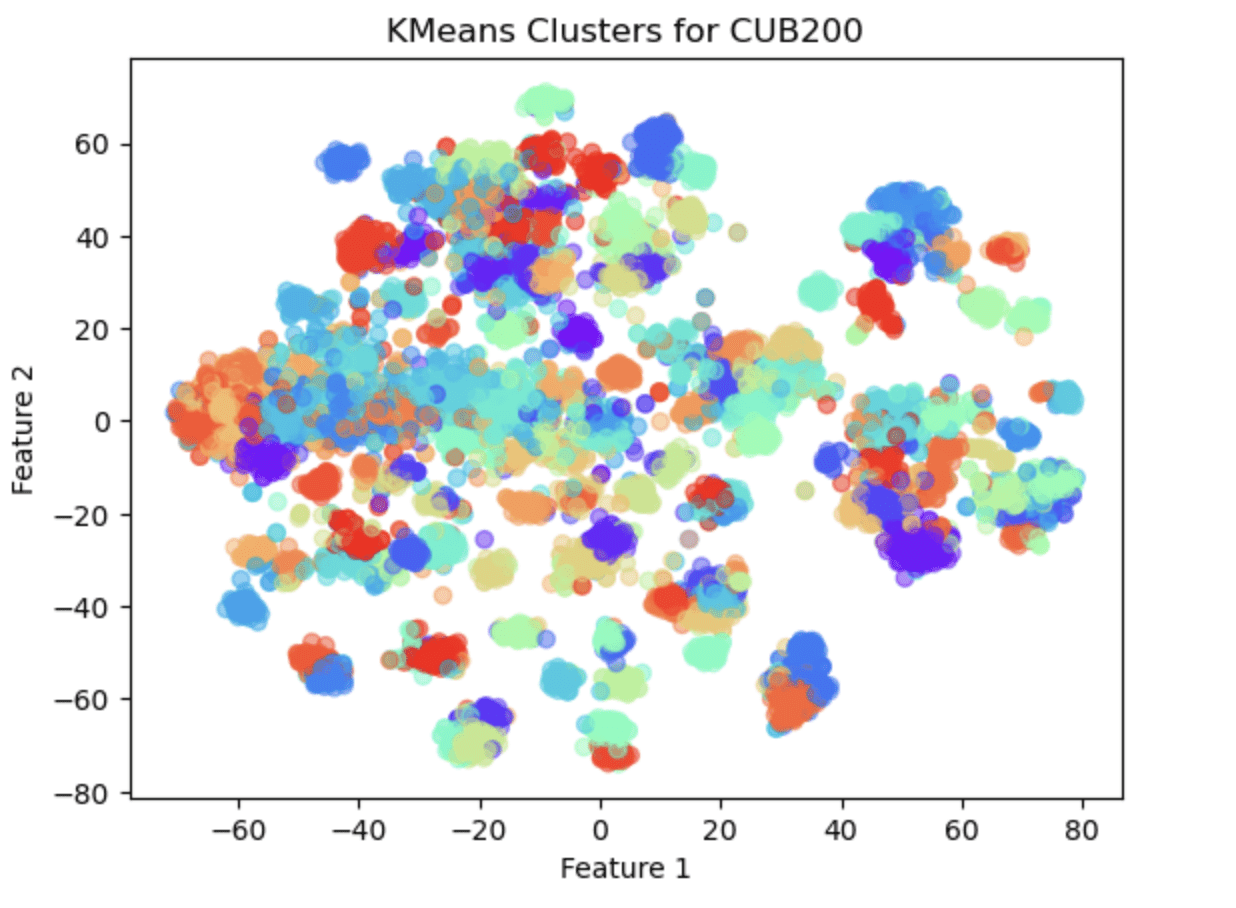}} 
\caption{Visualization of k-means clusters of CUB200 images with CLIP.}
\label{fig:kmeans}
\end{center}
\end{figure}

With k-means clustering method, we build a distinguishable by CLIP clusters (see \cref{fig:kmeans}).

A crucial thing around t-SNE visualization is that the latent space of CLIP-like models is strictly divided into two clusters: one corresponding to image embeddings, and the second one, which stands for textual modality. Intuition behind CBMs suggests that the distribution of modalities should be completely different from an observed in \cref{fig:tsne}. Indeed, for the sake of interpretability, embeddings of an image are expected to stay closer to the vectors of the corresponding concepts and classes. If it is true, then a simple kNN algorithm can find the most relevant concepts, which is incorrect in our case. Thus, we highlight an unresolved problem of building models that will learn this kind of latent space in end-to-end manner.

\subsection{Concepts Interpretability}
\label{sec:concepts_interpretability}
\begin{figure}[h] 
\centering
   \begin{subfigure}
     \centering
    \includegraphics[width=0.75\linewidth]{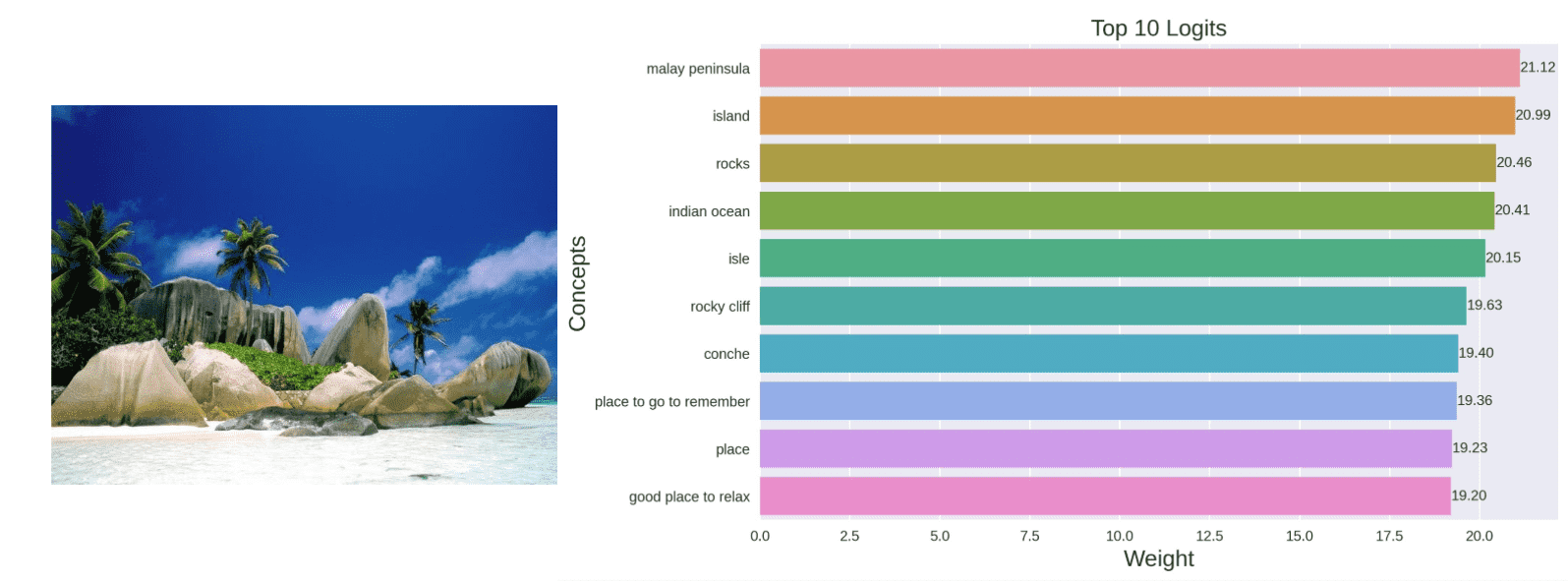}
    \caption{Concepts extracted by CLIP.}
    \label{fig:clip_im_1}
    \end{subfigure}
    \begin{subfigure}
    \centering
      \includegraphics[width=0.75\linewidth]{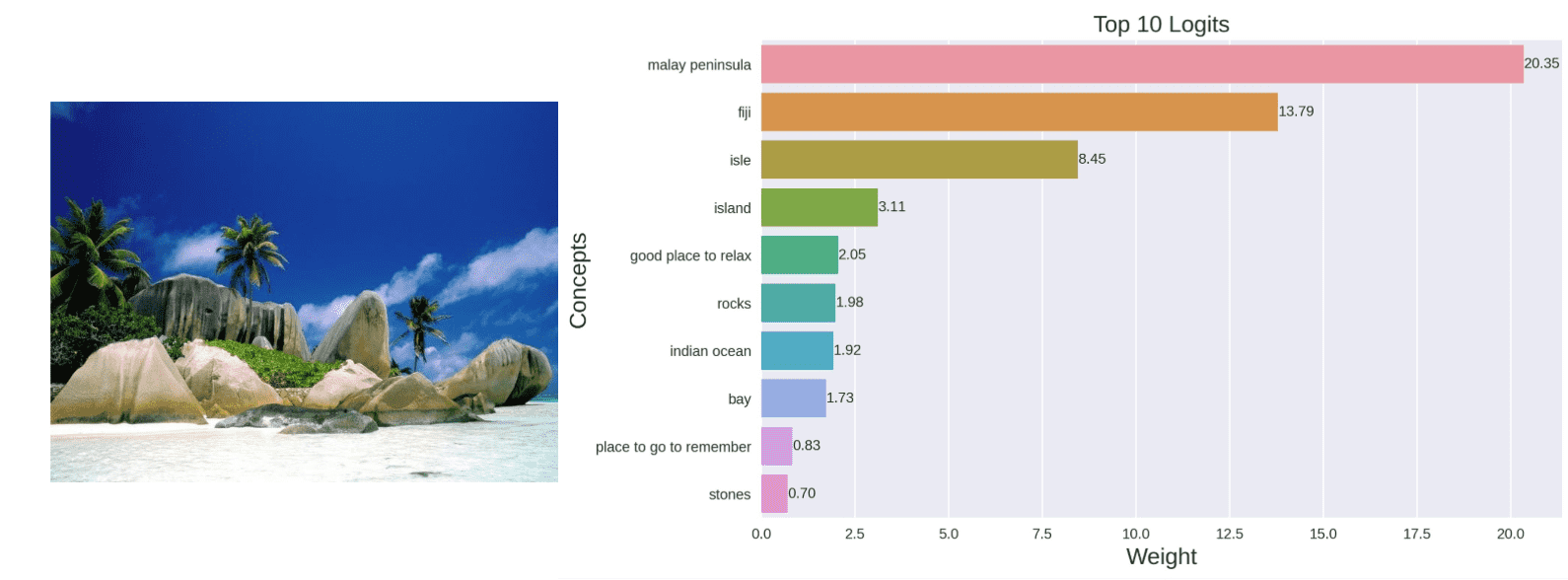}
    \caption{Concepts extracted by Sparse-CBM.}
    \label{fig:sparse_im_1}
    \end{subfigure}
    \begin{subfigure}
     \centering
  \includegraphics[width=0.75\linewidth]{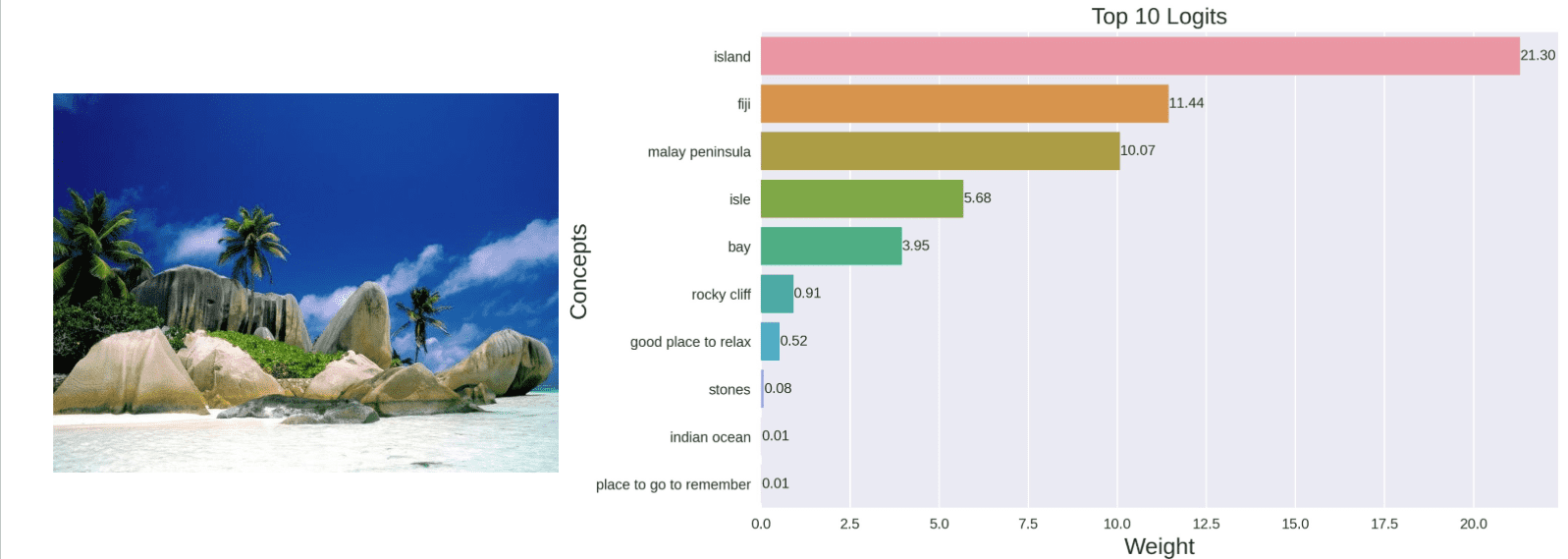}
    \caption{Concepts extracted by $\ell_1$-CBM.}
    \label{fig:l1_im_1}
    \end{subfigure}
        \begin{subfigure}
     \centering
  \includegraphics[width=0.75\linewidth]{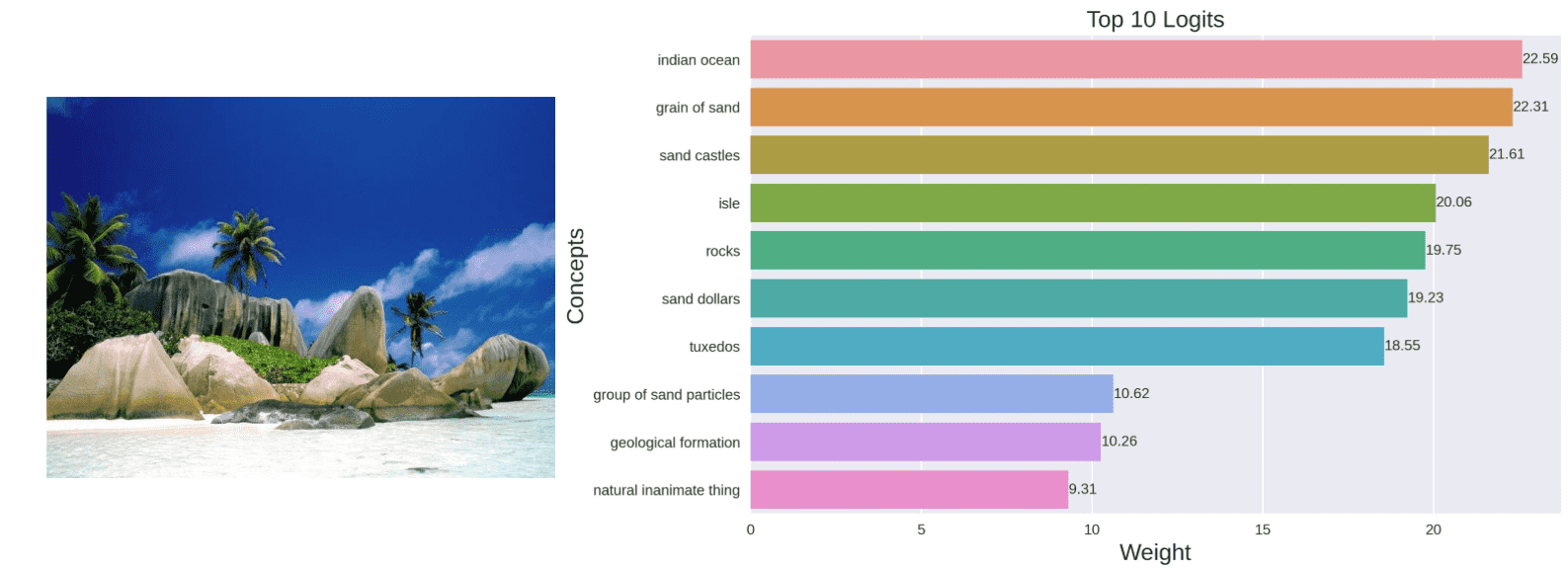}
    \caption{Concepts extracted by Contrastive-CBM.}
    \label{fig:contr_im_1}
    \end{subfigure}
    \caption{Concepts extracted by models trained on Places365.}
    \label{fig:concepts_1}
\end{figure}

\begin{figure}[h] 
\centering
   \begin{subfigure}
     \centering
    \includegraphics[width=0.75\linewidth]{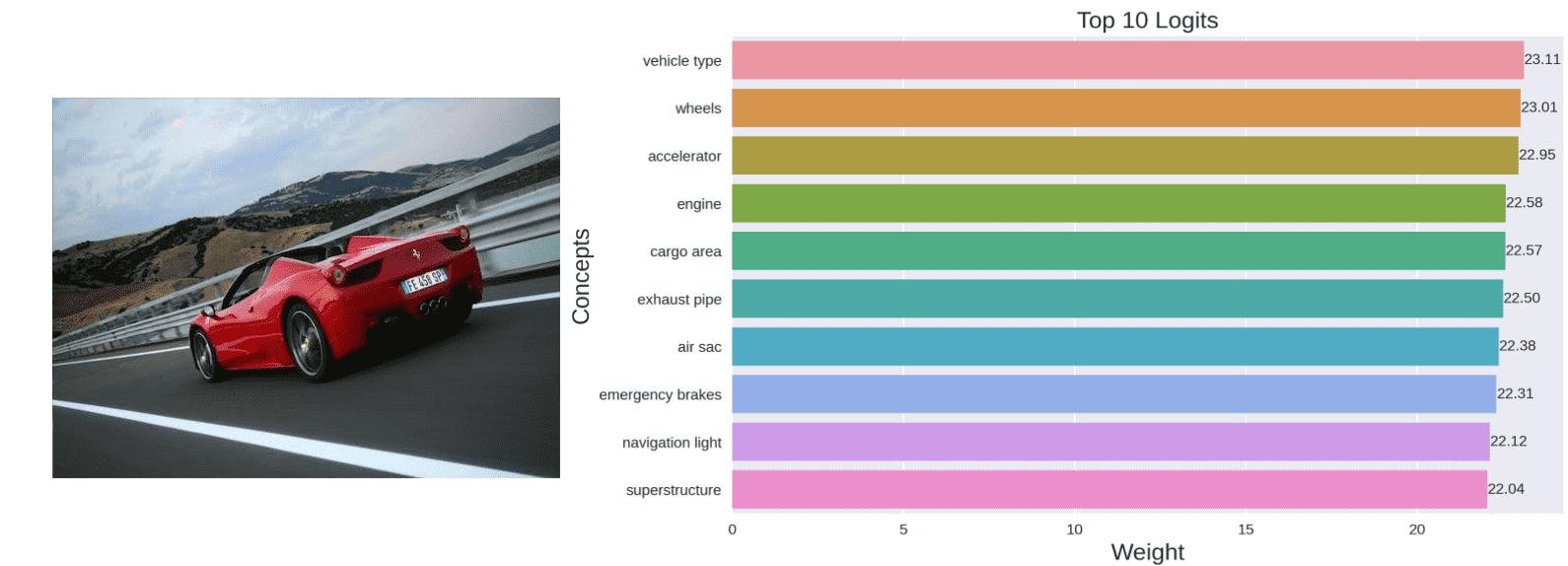}
    \caption{Concepts extracted by CLIP.}
    \label{fig:clip_im_2}
    \end{subfigure}
    \begin{subfigure}
    \centering
      \includegraphics[width=0.75\linewidth]{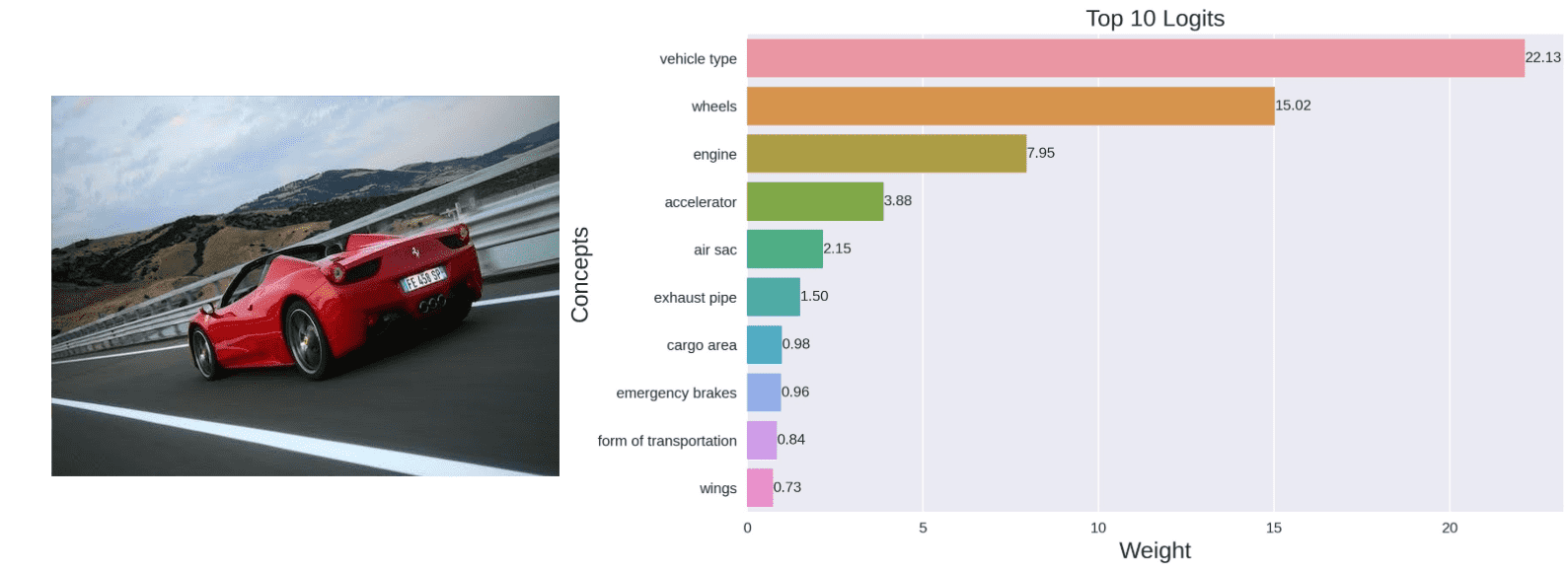}
    \caption{Concepts extracted by Sparse-CBM.}
    \label{fig:sparse_im_2}
    \end{subfigure}
    \begin{subfigure}
     \centering
  \includegraphics[width=0.75\linewidth]{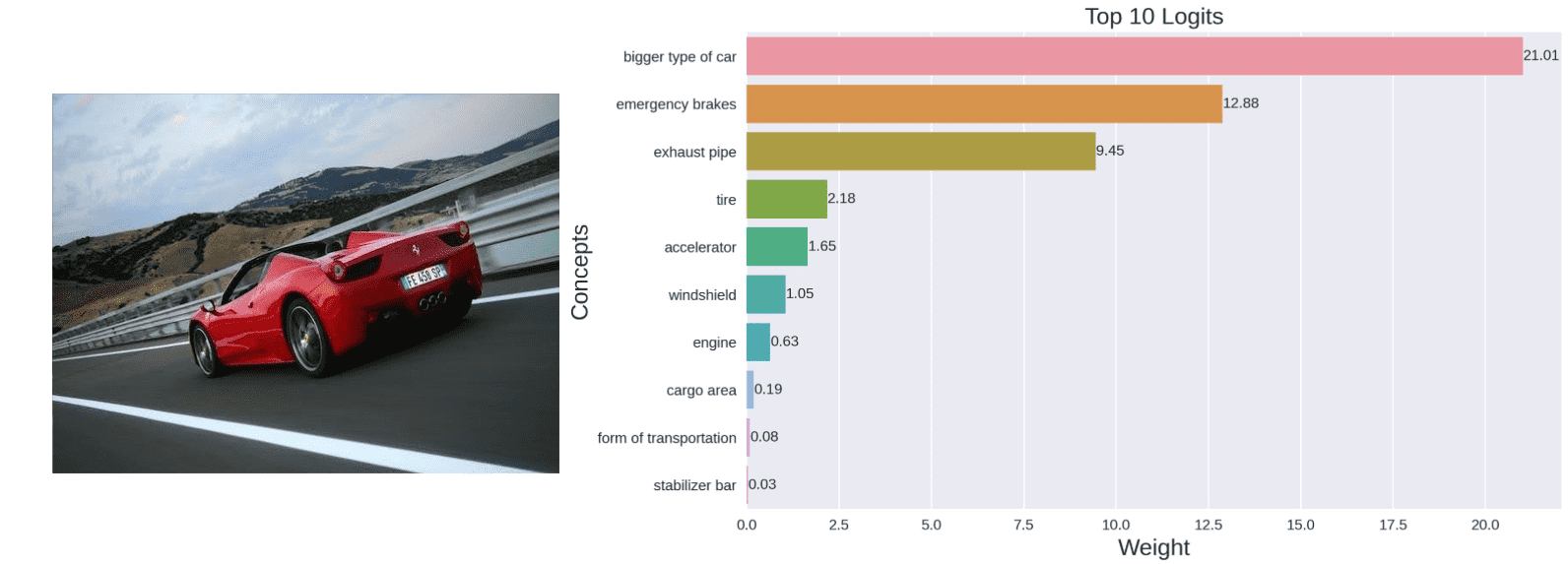}
    \caption{Concepts extracted by $\ell_1$-CBM.}
    \label{fig:l1_im_2}
    \end{subfigure}
        \begin{subfigure}
     \centering
  \includegraphics[width=0.75\linewidth]{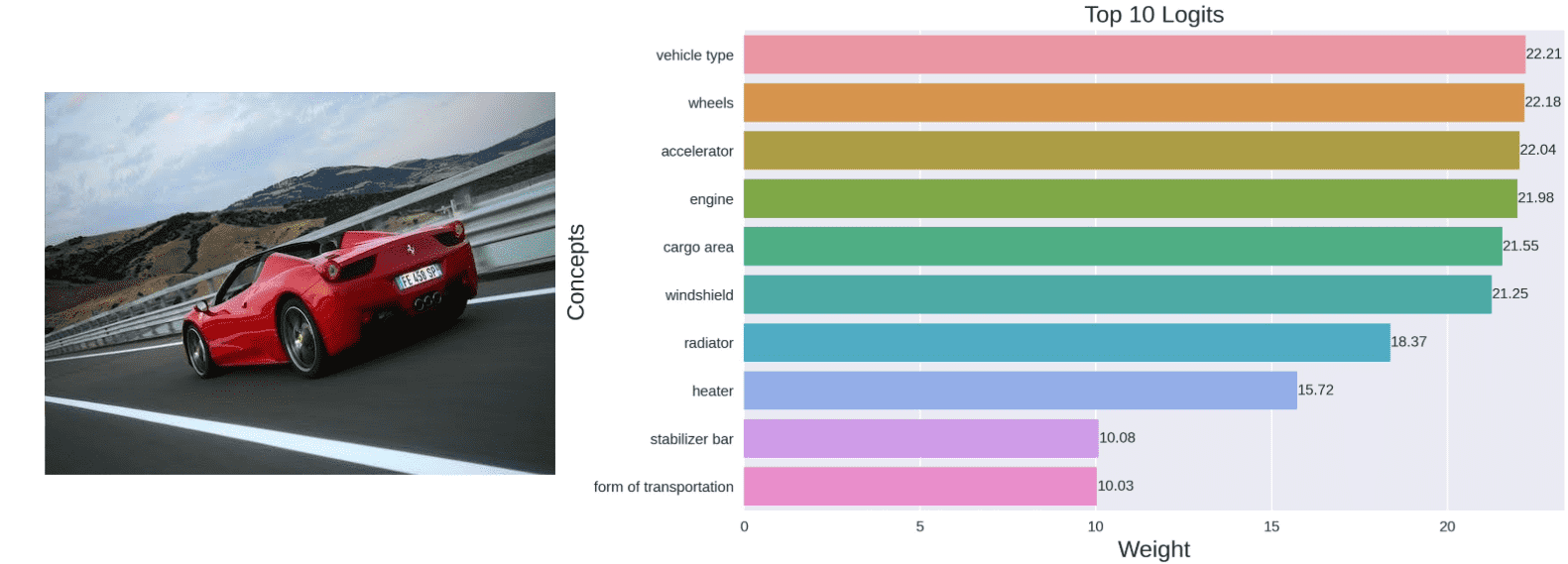}
    \caption{Concepts extracted by Contrastive-CBM.}
    \label{fig:contr_im_2}
    \end{subfigure}
    \caption{Concepts extracted by models trained on CIFAR10.}
    \label{fig:concepts_2}
\end{figure}

\begin{figure}[h] 
\centering
   \begin{subfigure}
     \centering
    \includegraphics[width=0.75\linewidth]{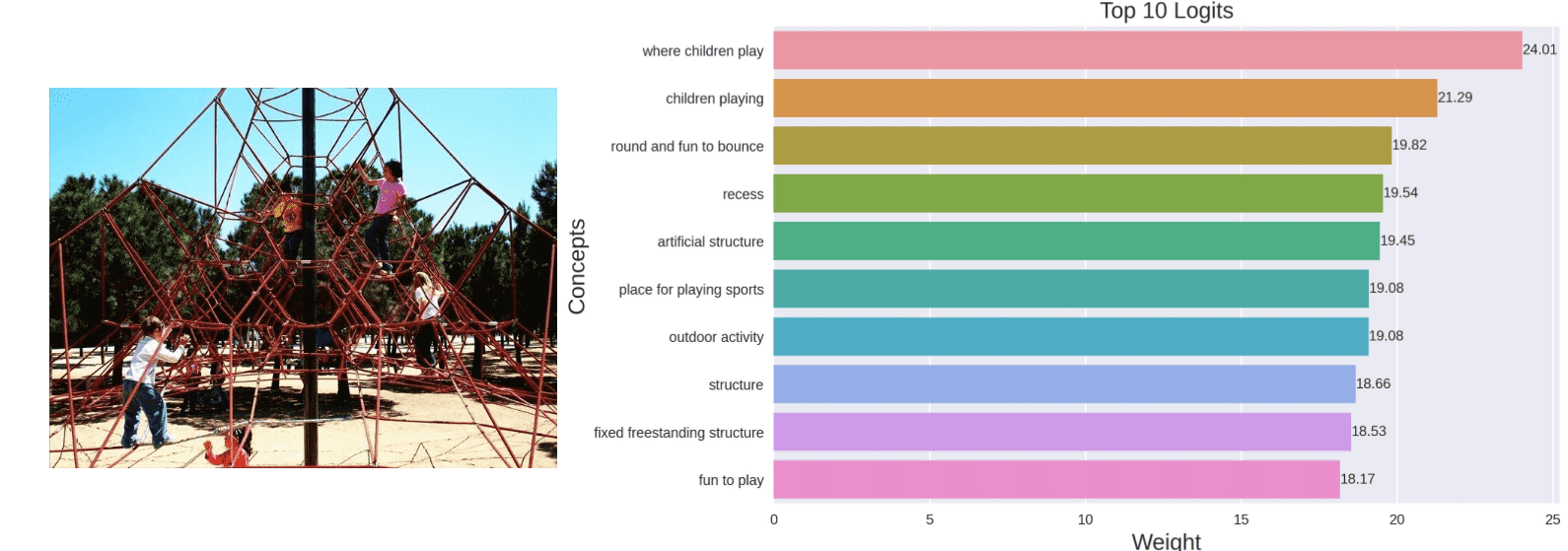}
    \caption{Concepts extracted by CLIP.}
    \label{fig:clip_im_3}
    \end{subfigure}
    \begin{subfigure}
    \centering
      \includegraphics[width=0.75\linewidth]{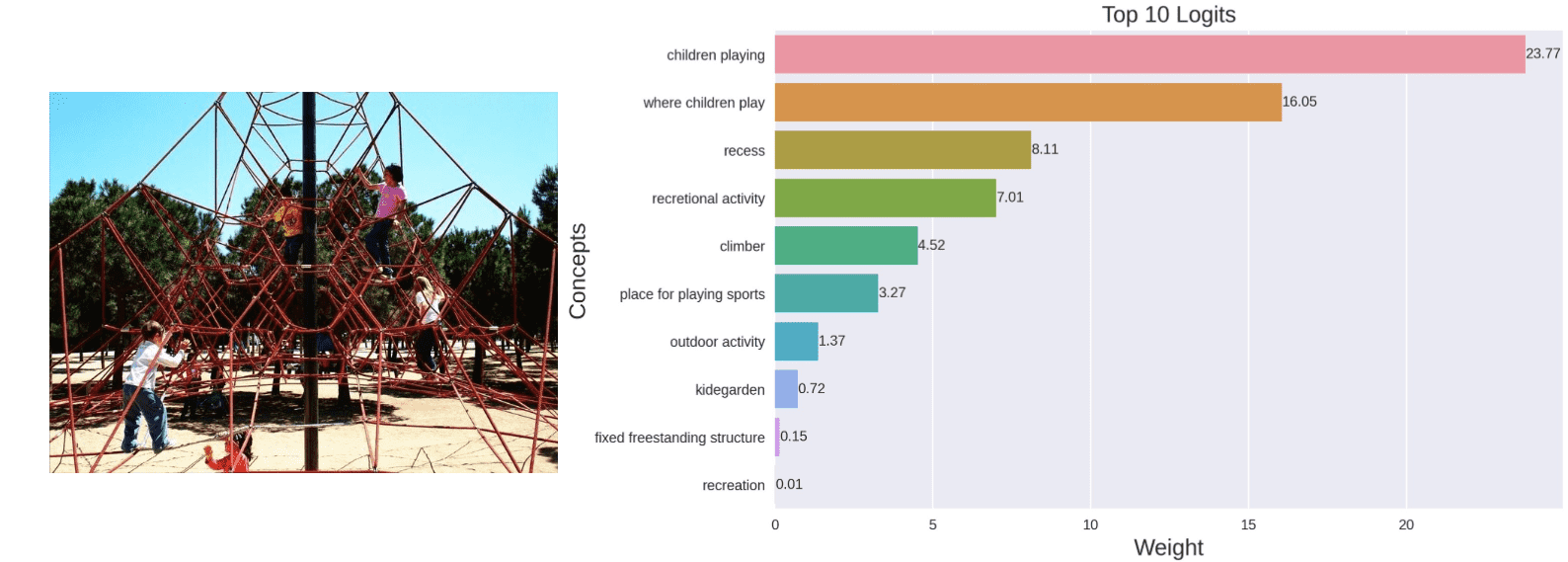}
    \caption{Concepts extracted by Sparse-CBM.}
    \label{fig:sparse_im_3}
    \end{subfigure}
    \begin{subfigure}
     \centering
  \includegraphics[width=0.75\linewidth]{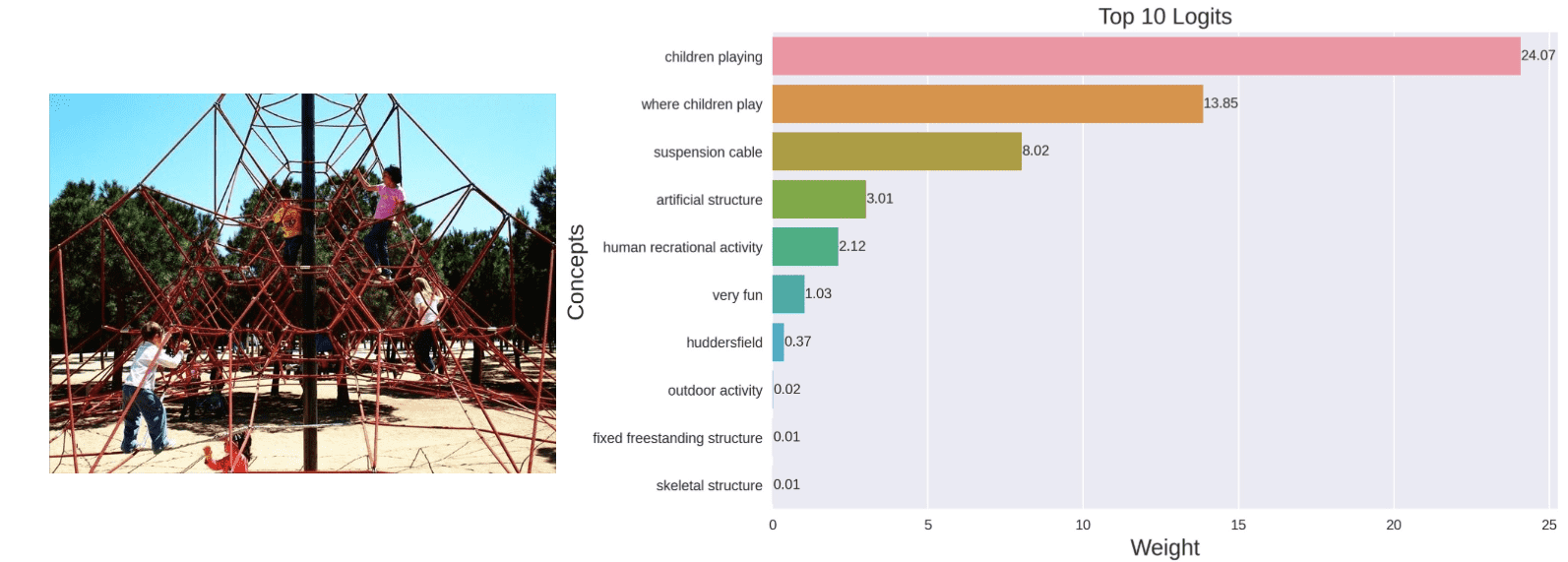}
    \caption{Concepts extracted by $\ell_1$-CBM.}
    \label{fig:l1_im_3}
    \end{subfigure}
        \begin{subfigure}
     \centering
  \includegraphics[width=0.75\linewidth]{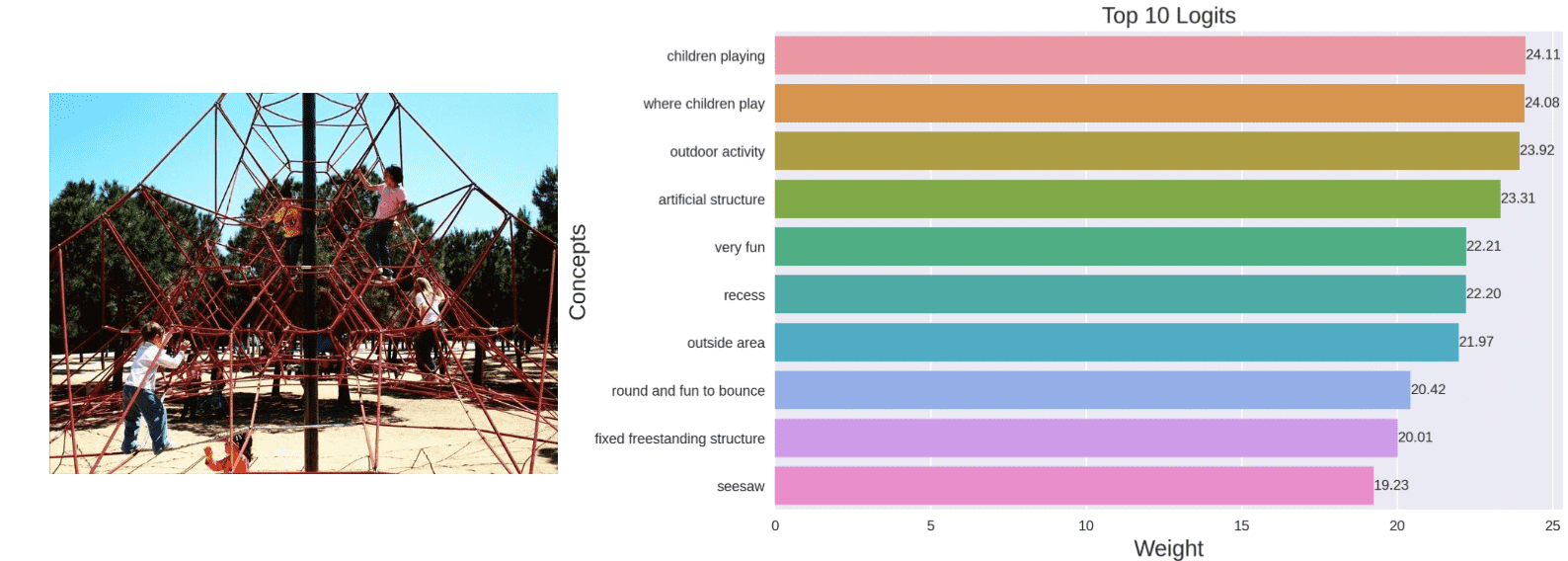}
    \caption{Concepts extracted by Contrastive-CBM.}
    \label{fig:contr_im_3}
    \end{subfigure}
    \caption{Concepts extracted by models trained on Places365.}
    \label{fig:concepts_3}
\end{figure}

\begin{figure}[h] 
\centering
   \begin{subfigure}
     \centering
    \includegraphics[width=0.65\linewidth]{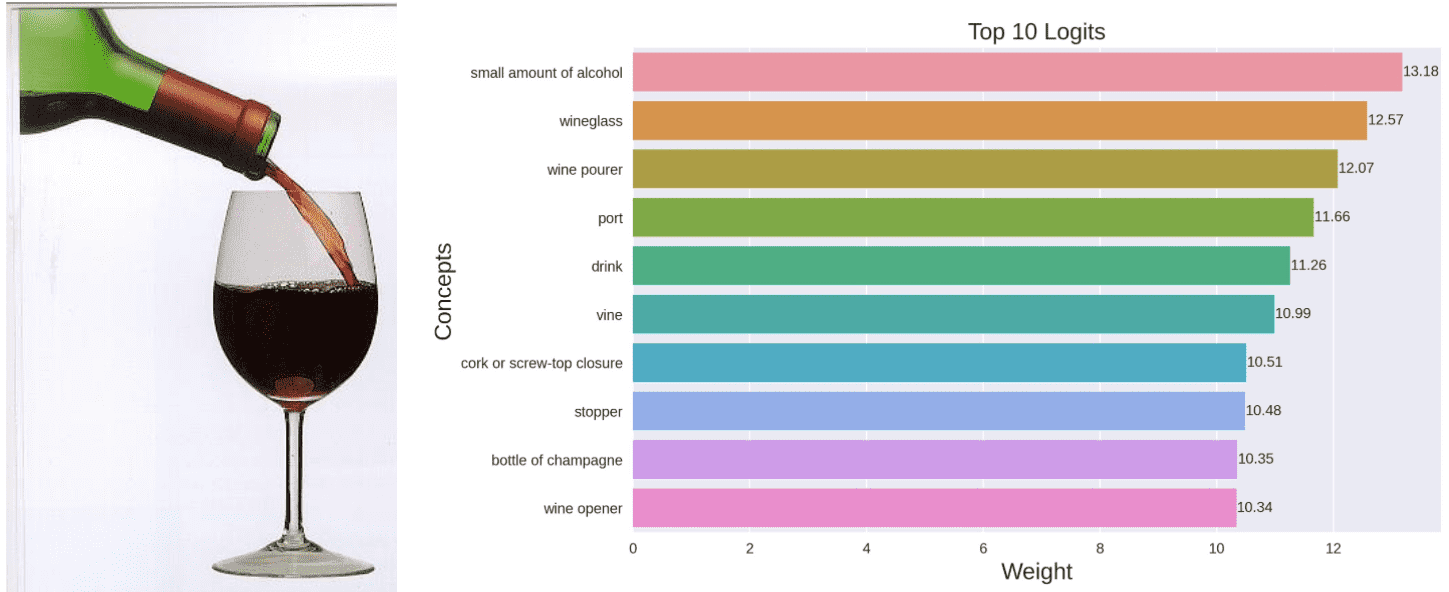}
    \caption{Concepts extracted by CLIP.}
    \label{fig:clip_im_4}
    \end{subfigure}
    \begin{subfigure}
    \centering
      \includegraphics[width=0.65\linewidth]{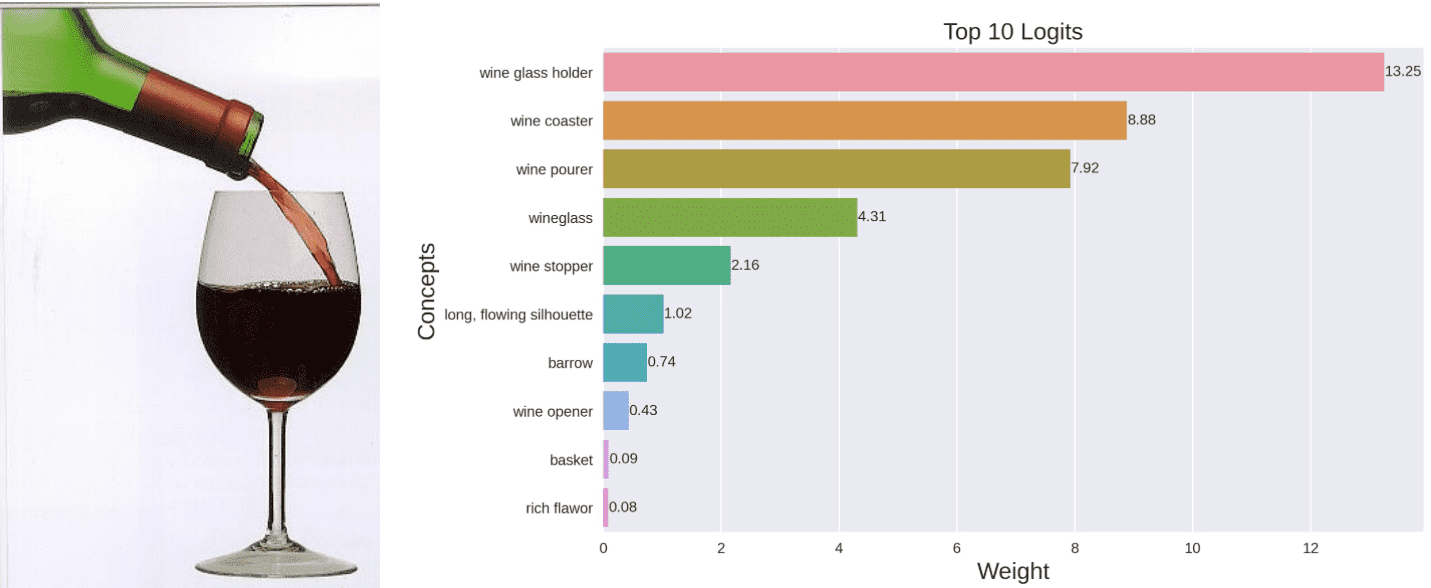}
    \caption{Concepts extracted by Sparse-CBM.}
    \label{fig:sparse_im_4}
    \end{subfigure}
    \begin{subfigure}
     \centering
  \includegraphics[width=0.65\linewidth]{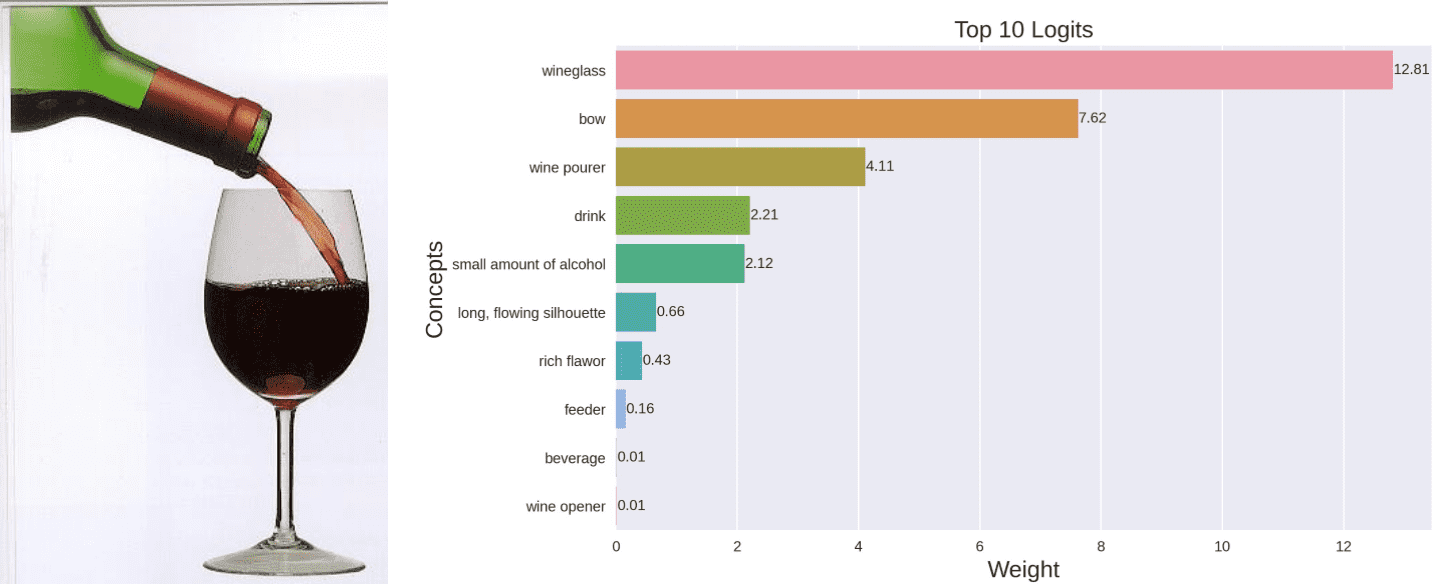}
    \caption{Concepts extracted by $\ell_1$-CBM.}
    \label{fig:l1_im_4}
    \end{subfigure}
        \begin{subfigure}
     \centering
  \includegraphics[width=0.65\linewidth]{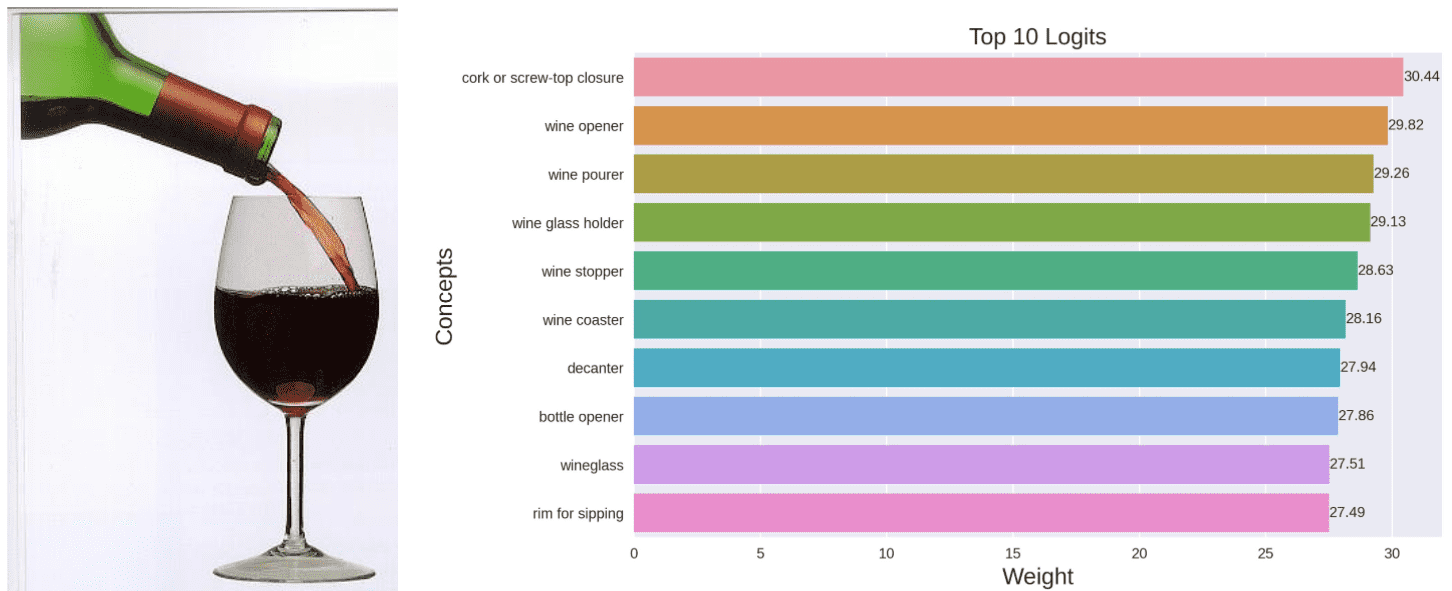}
    \caption{Concepts extracted by Contrastive-CBM.}
    \label{fig:contr_im_4}
    \end{subfigure}
    \caption{Concepts extracted by models trained on ImageNet.}
    \label{fig:concepts_4}
\end{figure}

\begin{figure}[h] 
\centering
   \begin{subfigure}
     \centering
    \includegraphics[width=0.7\linewidth]{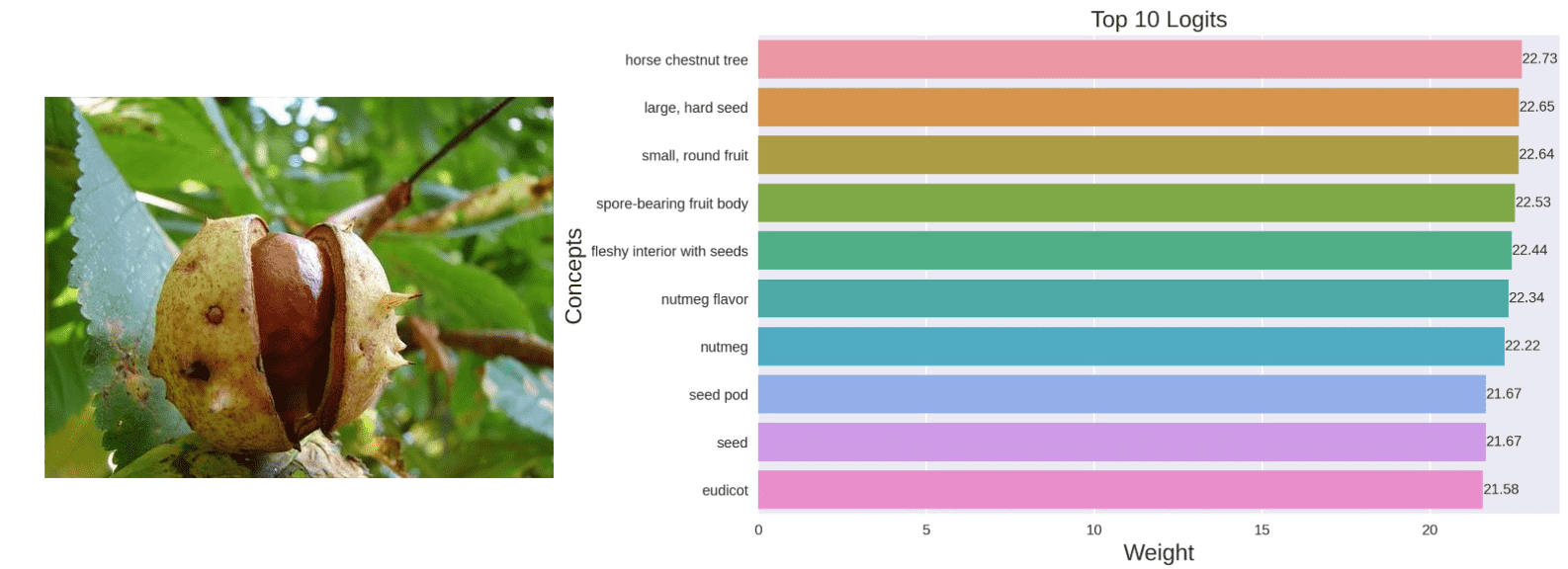}
    \caption{Concepts extracted by CLIP.}
    \label{fig:clip_im_5}
    \end{subfigure}
    \begin{subfigure}
    \centering
      \includegraphics[width=0.7\linewidth]{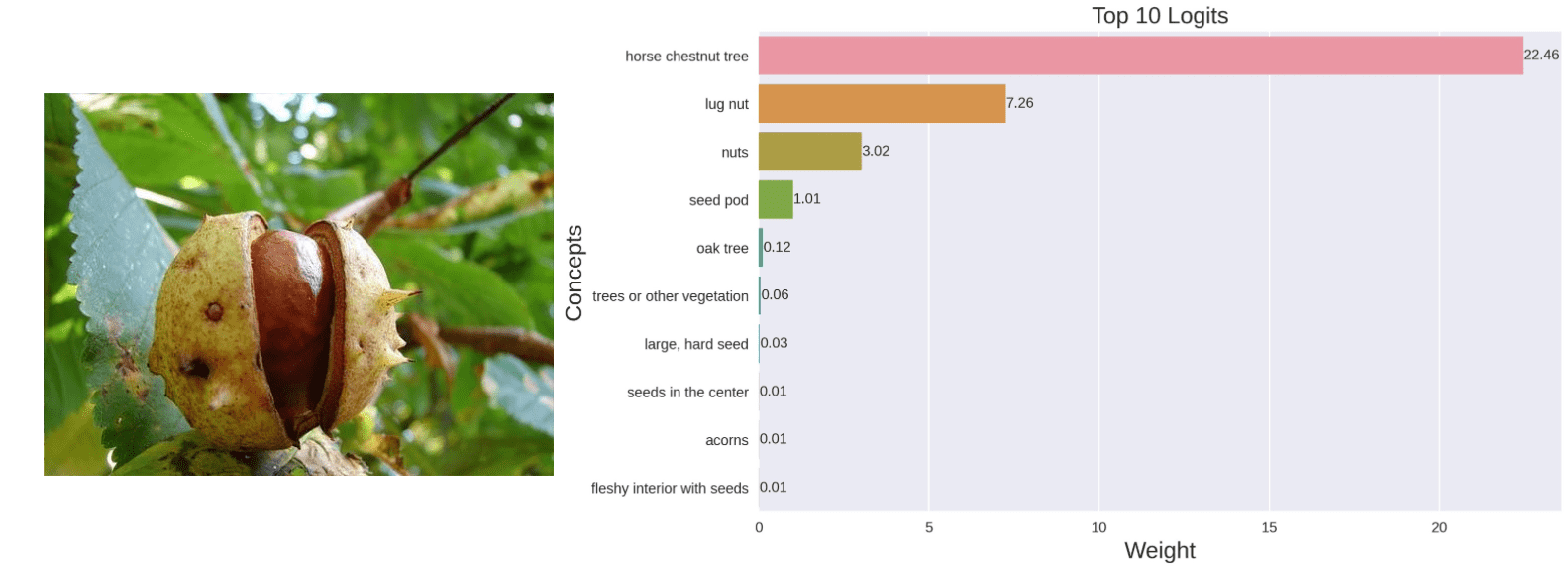}
    \caption{Concepts extracted by Sparse-CBM.}
    \label{fig:sparse_im_5}
    \end{subfigure}
    \begin{subfigure}
     \centering
  \includegraphics[width=0.75\linewidth]{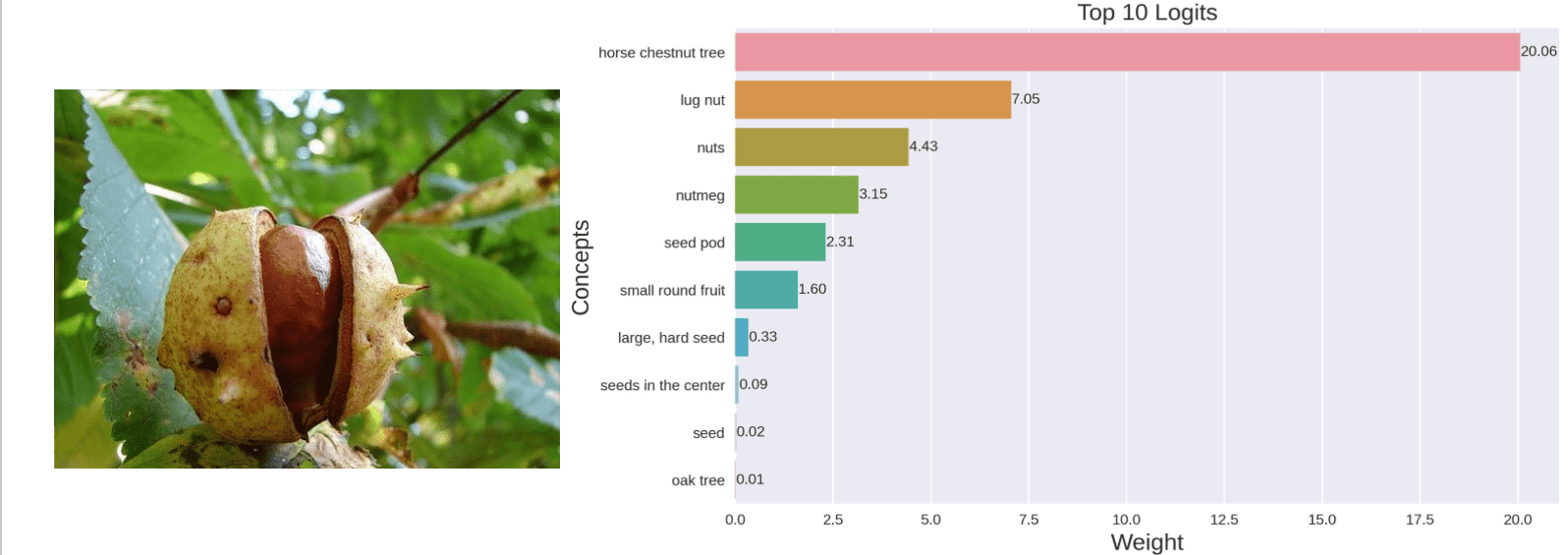}
    \caption{Concepts extracted by $\ell_1$-CBM.}
    \label{fig:l1_im_5}
    \end{subfigure}
        \begin{subfigure}
     \centering
  \includegraphics[width=0.75\linewidth]{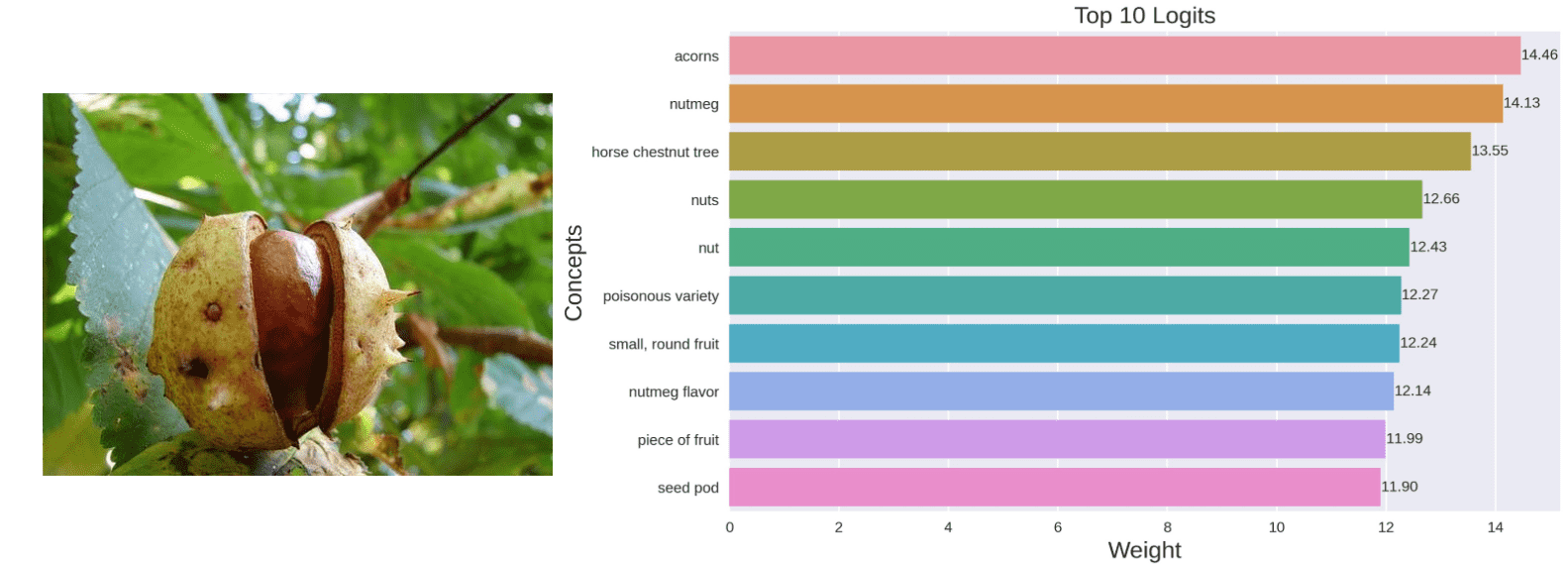}
    \caption{Concepts extracted by Contrastive-CBM.}
    \label{fig:contr_im_5}
    \end{subfigure}
    \caption{Concepts extracted by models trained on Places365.}
    \label{fig:concepts_5}
\end{figure}

\begin{figure}[h] 
\centering
   \begin{subfigure}
     \centering
    \includegraphics[width=0.75\linewidth]{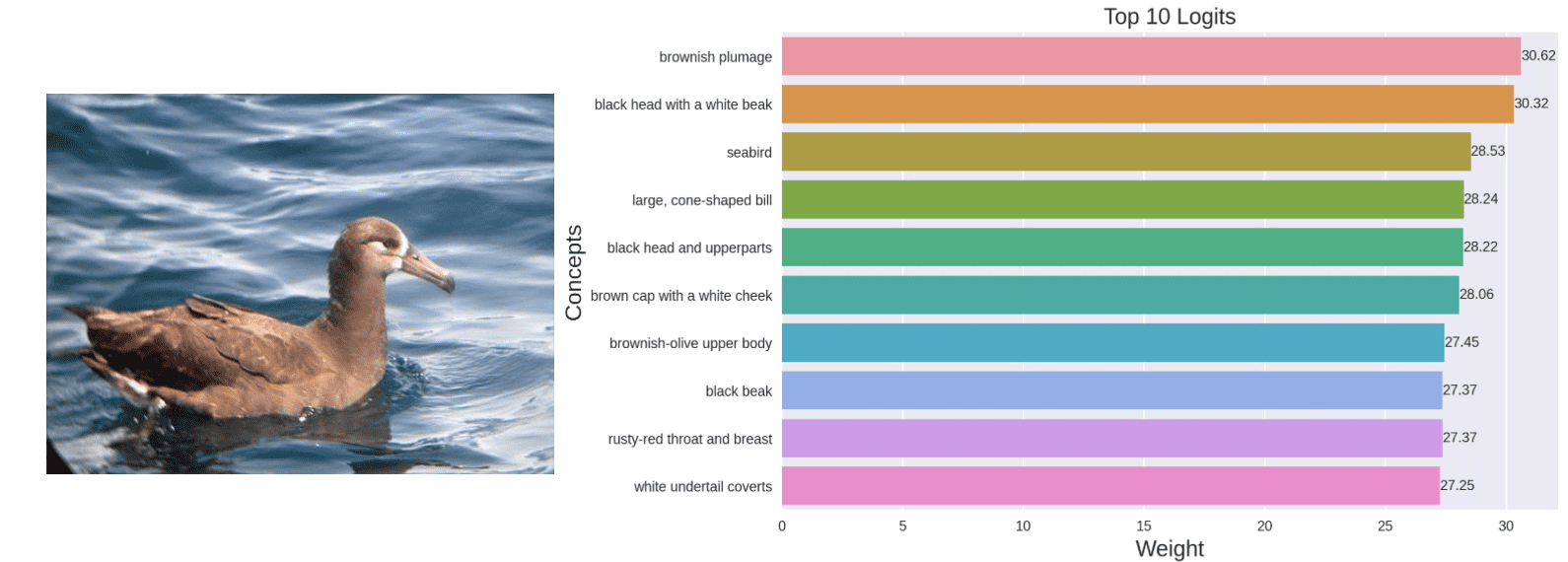}
    \caption{Concepts extracted by CLIP.}
    \label{fig:clip_im_6}
    \end{subfigure}
    \begin{subfigure}
    \centering
      \includegraphics[width=0.75\linewidth]{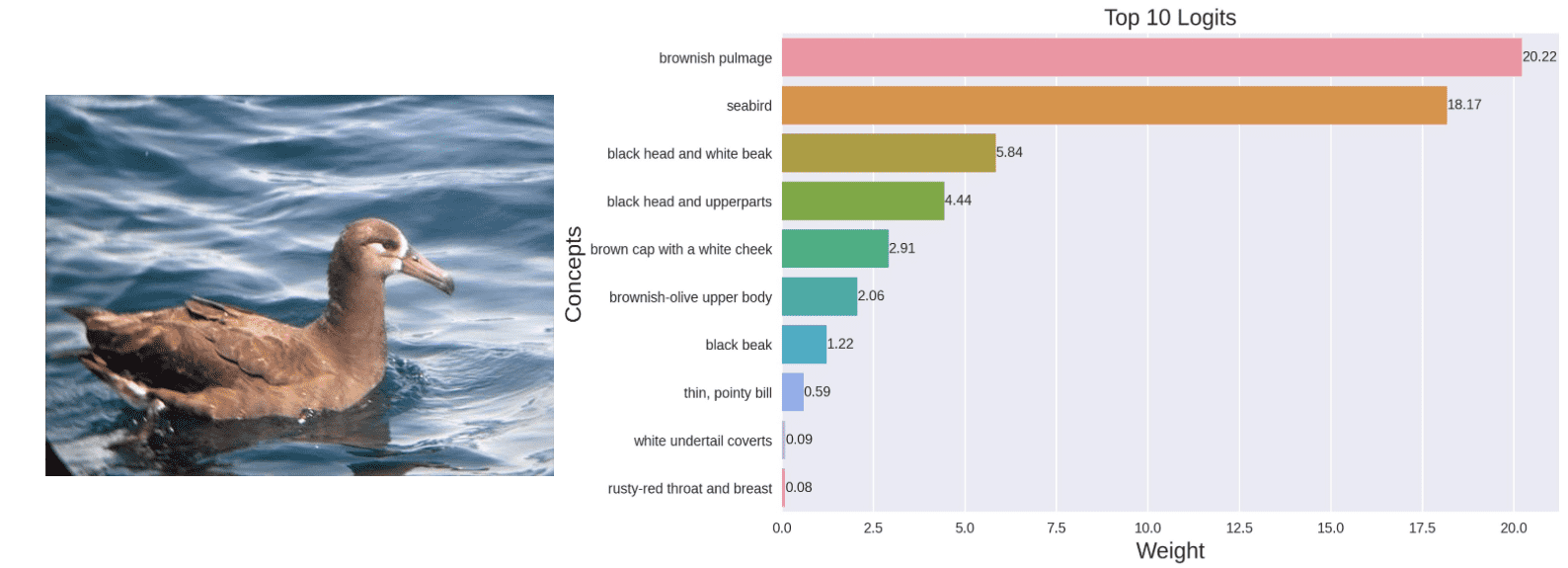}
    \caption{Concepts extracted by Sparse-CBM.}
    \label{fig:sparse_im_6}
    \end{subfigure}
    \begin{subfigure}
     \centering
  \includegraphics[width=0.75\linewidth]{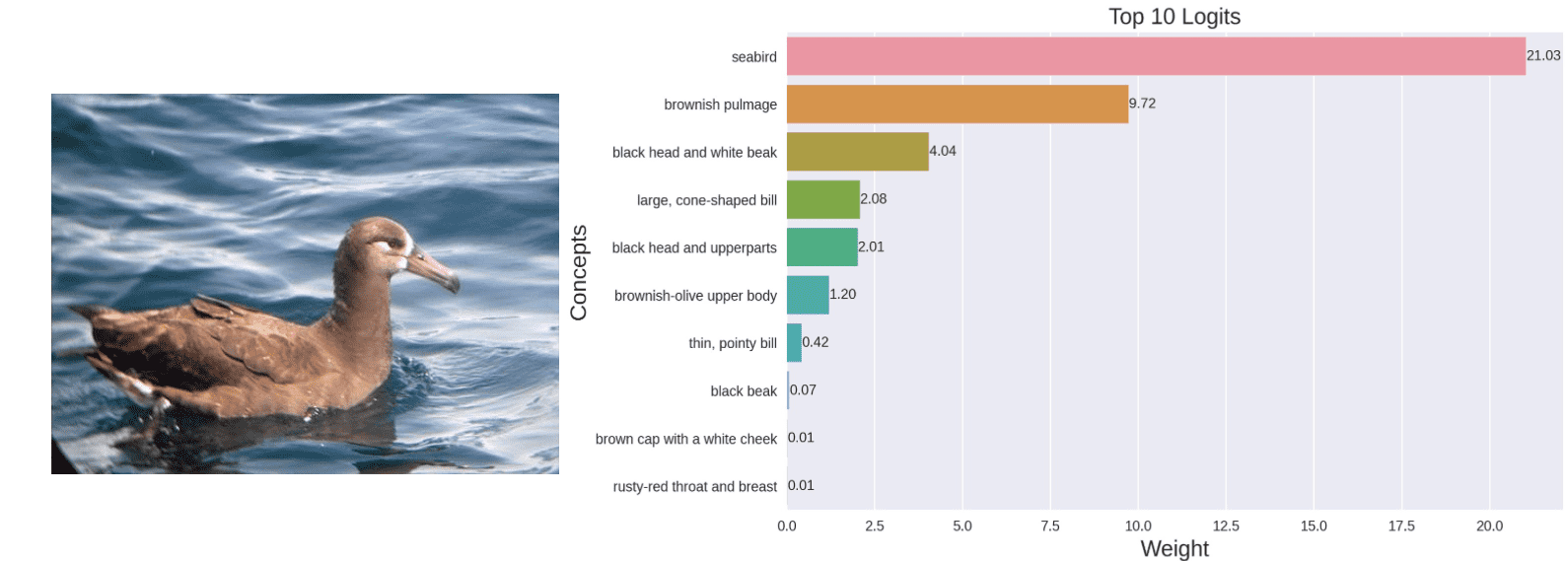}
    \caption{Concepts extracted by $\ell_1$-CBM.}
    \label{fig:l1_im_6}
    \end{subfigure}
        \begin{subfigure}
     \centering
  \includegraphics[width=0.75\linewidth]{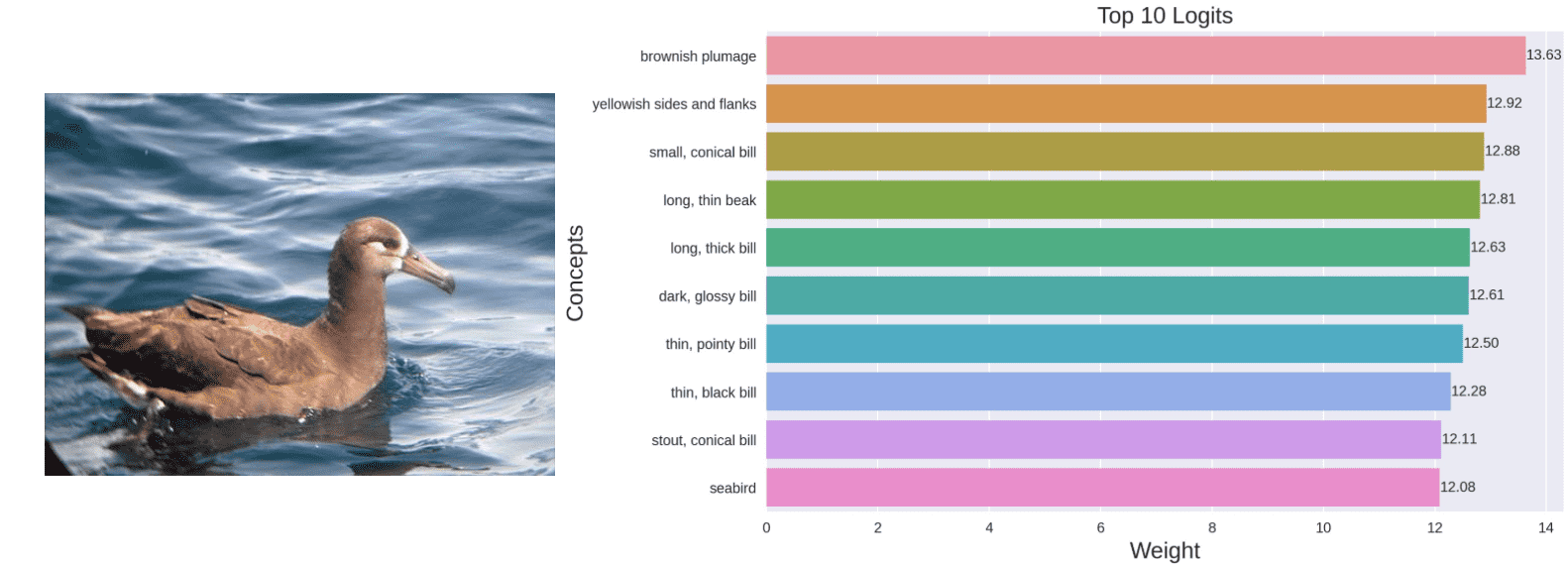}
    \caption{Concepts extracted by Contrastive-CBM.}
    \label{fig:contr_im_6}
    \end{subfigure}
    \caption{Concepts extracted by models trained on CUB200.}
    \label{fig:concepts_6}
\end{figure}

We compare an interpretability power of frawework with basic CLIP feature extracting properties on the subset of images from Places365, CUB200, CIFAR10 and ImageNet datasets. We select both basic images, for clarity, and the complex ones with many inner details. For CLIP-like models, we record the top-k (k=10) highest dot-product scores, while for our framework architectures we look at Concept Bottleneck Layer outputs. We note that further results in \cref{fig:concepts_1,fig:concepts_2,fig:concepts_3,fig:concepts_4,fig:concepts_5,fig:concepts_6} are conducted using CLIP-ViT-L/14 backbone model. We definitely observe a sparsified activations with both of Sparse- and $\ell_1$-CBMs. And show that Concept Bottleneck Layer trained with contrastive objective produce a quite similar activation compared to backbone CLIP. At the same time, the final accuracy becomes higher with Sparse and $\ell_1$ models which claims a superiority of methods that sparsify the inner layers of model.

In this work we do not present an interpretability bars of Concept Matrix Search because this approach to classification with concept bottleneck does not modify the backbone CLIP model, thus, it has the same properties shown in \cref{fig:clip_im_1,fig:clip_im_2,fig:clip_im_3,fig:clip_im_4,fig:clip_im_5,fig:clip_im_6}.

\subsection{Sparse-CBM confusion}\label{sec:sparse_confusion}
\begin{figure}[t]
\begin{center}
\centerline{
\includegraphics[width=0.6\columnwidth]{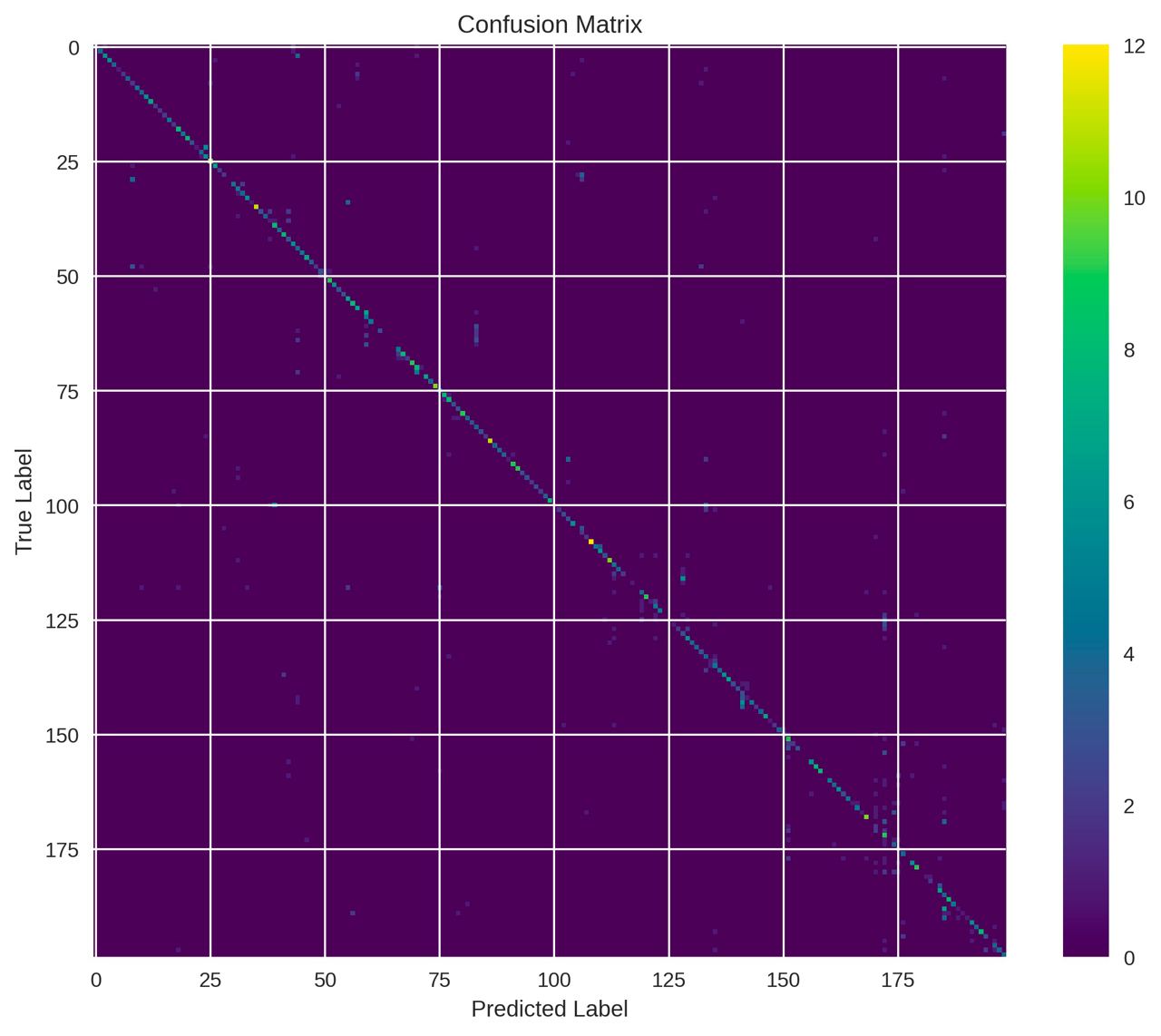}}
\caption{Confusion matrix of Sparse-CBM predictions on CUB200 dataset.}
\label{fig:cub_conf_matrix}
\end{center}
\end{figure}

Along with the final performance in classification, we report a confusion matrix of the best Sparse-CBM trained on CUB200 dataset which achieves 80.02\% accuracy in \cref{fig:cub_conf_matrix}. One should observe, that the most significant model errors are laying in the latter part of labels. This includes several similar classes such as the "Black capped Vireo", "Blue headed Vireo", "Philadelphia Vireo", "Red eyed Vireo", "Warbling Vireo", "White eyed Vireo" and "Yellow throated Vireo". All in all, in diverse datasets like CUB200, many similar classes are presented, therefore, it is also meaningful to consider concept extraction and to highlight their differences between classes like "Red eyed Vireo" and "White eyed Vireo".

\end{document}